\documentclass[10pt,journal,compsoc]{IEEEtran}
\usepackage{afterpage}
\usepackage{xcolor} 
\usepackage{calc}   
\newlength{\myorigtextheight}

\usepackage{amsmath,amssymb,amsfonts,amsthm}
\usepackage{algorithm}
\usepackage{algorithmic}
\usepackage{cite}
\usepackage{color}
\usepackage{graphicx}
\usepackage{hyperref}
\usepackage{booktabs}
\usepackage{multirow}
\usepackage{fancyhdr}
\usepackage{hyperref}
\usepackage{tikz}
\usepackage{fancyvrb} 
\usepackage[most]{tcolorbox}  
\tcbuselibrary{listingsutf8}
\usepackage{listings}
\usepackage{courier} 
\fancypagestyle{IEEEtitlepagestyle}{
  \fancyhf{}
  \fancyhead[L]{Please refer to the journal version:}
  \fancyhead[R]{\href{https://doi.org/10.1109/TPAMI.2025.3537087}{IEEE TPAMI 2025  DOI:10.1109/TPAMI.2025.3537087}}
}

\newcommand*\bibkey[1]{\textcolor{black!60}{\normalfont\bfseries\makebox[\widthof{\textbf{journal}}][l]{#1}}}
\newcommand*\bibval[1]{ #1}

\newcommand\copyrighttext{%
  \scriptsize
  Please refer to the journal version:
  
  \parbox{\linewidth}{%
    \bibval{@article\{duan2025dsact,}\\
    \hspace*{1em}\bibkey{title} = \bibval{\{Distributional Soft Actor-Critic with Three Refinements\},}\\
        \hspace*{1em}\bibkey{author} = \bibval{\{Duan, J. and Wang, W. and Xiao, L. and Gao, J. and Li, S. E. and Liu, C. and Zhang, Y.-Q. and Cheng, B. and Li, K.\},}\\
    \hspace*{1em}\bibkey{journal} = \bibval{\{IEEE Transactions on Pattern Analysis and Machine Intelligence\},}\\
    \hspace*{1em}\bibkey{volume} = \bibval{\{47\},}\\
    \hspace*{1em}\bibkey{number} = \bibval{\{5\},}\\
    \hspace*{1em}\bibkey{pages} = \bibval{\{3935--3946\},}\\
    \hspace*{1em}\bibkey{year} = \bibval{\{2025\},}\\
    \hspace*{1em}\bibkey{doi} = \bibval{\{10.1109/TPAMI.2025.3537087\}}\\
    \bibval{\}}%
  }
}

\newcommand\copyrightnotice{%
\begin{tikzpicture}[remember picture,overlay]
\node[anchor=south,yshift=1pt] at (current page.south) {\fbox{\parbox{\dimexpr\textwidth-\fboxsep-\fboxrule\relax}{\copyrighttext}}};
\end{tikzpicture}%
}

\newtcolorbox{ieeebibbox}{
  colback=gray!3, colframe=gray!80,
  fonttitle=\bfseries,
  title=Official Citation (BibTeX),
  boxrule=0.4pt,
  sharp corners,
  width=\textwidth,
  left=5pt, right=5pt, top=5pt, bottom=5pt
}

\ifCLASSOPTIONcompsoc \usepackage[caption=false,font=normalsize,labelfont=sf,textfont=sf]{subfig}
\else
\usepackage[caption=false,font=footnotesize]{subfig}
\fi
\hypersetup{hidelinks,
	colorlinks=true,
	allcolors=black,
	pdfstartview=Fit,
	breaklinks=true}

\newcommand{\Exp}{\mathop{\mathbb E}\displaylimits}

\begin{document}

\title{Distributional Soft Actor-Critic with Three Refinements}

\author{Jingliang Duan,
        Wenxuan Wang, Liming Xiao, Jiaxin Gao, Shengbo Eben Li, Chang Liu, Ya-Qin Zhang, Bo Cheng, Keqiang Li
\IEEEcompsocitemizethanks{\IEEEcompsocthanksitem J. Duan is with the School of Mechanical Engineering, University of Science and Technology Beijing, Beijing 100083, China. He was with the School of Vehicle and Mobility, Tsinghua University, Beijing 100084, China (e-mail: duanjl@ustb.edu.cn).
\IEEEcompsocthanksitem L. Xiao is with the School of Mechanical Engineering, University of Science and Technology Beijing, Beijing 100083, China (e-mail: xlming@xs.ustb.edu.cn).
\IEEEcompsocthanksitem W. Wang, J. Gao, B. Cheng, and K. Li are with the School of Vehicle and Mobility, Tsinghua University, Beijing 100084, China (e-mail: wwx.sora.gmail.com; gaojiaxin@mail.tsinghua.edu.cn; chengbo@tsing-hua.edu.cn; likq@tsinghua.edu.cn).
\IEEEcompsocthanksitem S. E. Li is with School of Vehicle and Mobility \& College of AI, Tsinghua University, Beijing 100084, China (e-mail: lishbo@tsinghua.edu.cn).
\IEEEcompsocthanksitem C. Liu is with the College of Engineering, Peking University, Beijing 100871, China (e-mail: changliucoe@pku.edu.cn).
\IEEEcompsocthanksitem Y. Zhang is Institute for AI Industry Research, Tsinghua University, Beijing 100084, China (e-mail: zhangyaqin@air.tsinghua.edu.cn).}
\thanks{This study was supported in part by the NSF China under Grants 52202487, in part by National Key Research and Development Program of China under grant number 2022YFB2502901, in part by the Fundamental Research Funds for the Central Universities under Grant FRF-OT-23-02, and in part by the Young Elite Scientists Sponsorship Program by CAST under Grant 2023QNRC001.  J. Duan and W. Wang contribute equally to this work. All correspondences should be sent to S. Li with email: lisb04@gmail.com.}
}




\IEEEtitleabstractindextext{
\begin{abstract}

Reinforcement learning (RL) has shown remarkable success in solving complex decision-making and control tasks. However, many model-free RL algorithms experience performance degradation due to inaccurate value estimation, particularly the overestimation of Q-values, which can lead to suboptimal policies. To address this issue, we previously proposed the Distributional Soft Actor-Critic (DSAC or DSACv1), an off-policy RL algorithm that enhances value estimation accuracy by learning a continuous Gaussian value distribution. Despite its effectiveness, DSACv1 faces challenges such as training instability and sensitivity to reward scaling, caused by high variance in critic gradients due to return randomness. In this paper, we introduce three key refinements to DSACv1 to overcome these limitations and further improve Q-value estimation accuracy: expected value substitution, twin value distribution learning, and variance-based critic gradient adjustment. The enhanced algorithm, termed DSAC with Three refinements (DSAC-T or DSACv2), is systematically evaluated across a diverse set of benchmark tasks. Without the need for task-specific hyperparameter tuning, DSAC-T consistently matches or outperforms leading model-free RL algorithms, including SAC, TD3, DDPG, TRPO, and PPO, in all tested environments. Additionally, DSAC-T ensures a stable learning process and maintains robust performance across varying reward scales. Its effectiveness is further demonstrated through real-world application in controlling a wheeled robot, highlighting its potential for deployment in practical robotic tasks.
\end{abstract}

\begin{IEEEkeywords}
Reinforcement Learning, Dynamic Programming, Optimal Control, Neural Network
\end{IEEEkeywords}}

\addtolength{\textheight}{-2.0cm}
\maketitle

\thispagestyle{IEEEtitlepagestyle}
\IEEEdisplaynontitleabstractindextext
\IEEEpeerreviewmaketitle
\copyrightnotice
\afterpage{\setlength{\textheight}{\myorigtextheight}}

\IEEEraisesectionheading{\section{Introduction}}

\IEEEPARstart{R}{einforcement} learning (RL)  driven by the integration of high-capacity function approximators, such as neural networks, has succeeded in various demanding tasks, ranging from gaming scenarios to robotic control systems \cite{silver2016mastering,silver2017mastering,duan2024encoding,li2022decision,xiangkun_fear_2024}. Despite these achievements, typical model-free RL methods tend to learn unrealistically high state–action values. The convergence properties of RL, which hinge on accurate estimations of the value function \cite{duan2023optimization}, can be compromised when faced with inflated values, often resulting in significant performance declines. This problem is referred to as value overestimation, which is caused by the maximization operator of Bellman equation applied to noisy value estimates. Such a phenomenon is especially prevalent in deep RL setups due to inaccuracies in value function approximation, as seen in methods like deep Q-network (DQN) \cite{mnih2015DQN} and deep deterministic policy gradient (DDPG) \cite{lillicrap2015DDPG}. 

Double Q-learning, introduced by Hasselt \textit{et al.} \cite{hasselt2010double_Q}, is a pioneering method to mitigate overestimation. It achieves this ability by learning two distinct action-value functions, which decouple the maximization operation into action selection and action evaluation. A deep version of this approach, known as double DQN \cite{van2016double_DQN}, leverages the current Q-network to determine the greedy action, whereas its corresponding target Q-network is utilized to evaluate this action, resulting in a reduced estimation bias in comparison to traditional DQN methods \cite{mnih2015DQN}. One shortcoming of these two methods is that they are limited to handling only discrete action spaces. To deal with continuous action spaces, Fujimoto \textit{et al.} \cite{Fujimoto2018TD3} proposed clipped double Q-learning, an actor-critic-based approach that incorporates the minimum value between two Q-estimates in formulating the learning objective for both Q-value and policy functions. Since then, this scheme has been widely adopted in mainstream RL algorithms, including twin delayed deep deterministic policy gradient (TD3) \cite{Fujimoto2018TD3} and soft actor-critic (SAC) \cite{Haarnoja2018SAC,Haarnoja2018ASAC}, yielding substantial performance enhancements. Despite the success of these classical RL algorithms, the challenge of inaccurate value function estimation (whether it be overestimation or underestimation) continues to hinder further advancements in policy performance. In recent years, several novel methods have been developed to improve the accuracy of value function estimation \cite{sicen_realistic_2023, li_robust_2023, zhang_EAC_2024}. A notable example is the work by He \textit{et al.} \cite{xiangkun_trust_2024}, which augments the original reward function with a negative term representing the uncertainty of environmental dynamics. This approach enables safer and more robust policy evaluation by learning a confidence lower bound on the true Q-value.


The employment of a distributional value function is a new design trend for overestimation mitigation. In \cite{duan2021distributional}, we have introduced a standard version of the distributional soft actor-critic algorithm (also termed DSACv1) to enhance value estimation accuracy by learning a Gaussian distribution of random returns, referred to as the value distribution, rather than focusing solely on the expected return (i.e., Q-value). Mathematical analysis has shown that the overestimation bias is inversely proportional to the variance of the value distribution, which builds the theoretical foundation of overestimation mitigation caused by system randomness and approximation errors. However, the standard DSAC has certain limitations. One significant issue is occasional learning instability, which arises from the process of learning continuous Gaussian value distribution. This approach, which incorporates return randomness into the critic gradient, deviates from 
conventional RL that focuses solely on expected return, potentially leading to unstable critic updates. Another challenge is its sensitivity to reward scaling.  Standard DSAC employs a fixed boundary for clipping target returns, and the update magnitude of the Q-value is inversely related to the variance of the value distribution. Consequently, this design necessitates manual adjustments of the reward scale for different tasks to align with the fixed clipping boundary and to maintain a balanced Q-value update size.  Such a requirement for task-specific tuning limits the adaptability of DSAC across a range of tasks.

To address these issues of learning instability and reward scaling sensitivity, and to further enhance Q-value estimation accuracy, this study introduces an improved variant of DSAC, referred to as DSAC with three refinements (abbreviated as DSAC-T or DSACv2). This algorithm incorporates three key refinements to standard DSAC, including expected value substituting, twin value distribution learning, and variance-based critic gradient adjustment. Our empirical experiments show that DSAC-T outperforms or matches lots of mainstreaming model-free RL algorithms, including SAC, TD3, DDPG, TRPO, and PPO, across all benchmarked tasks. Moreover, compared with standard DSAC\cite{duan2021distributional}, DSAC-T exhibits better learning stability and diminishes the necessity for task-specific parameter adjustment. Interested readers can access the source code at \href{https://github.com/Jingliang-Duan/DSAC-T}{\emph{https://github.com/Jingliang-Duan/DSAC-v2}}. Additionally, DSAC-T is integrated into our self-developed open-source RL toolkit, named GOPS\footnote{Available at \href{https://github.com/Intelligent-Driving-Laboratory/GOPS}{\emph{https://github.com/Intelligent-Driving-Laboratory/GOPS}}} \cite{wang2023gops}.
The key contributions of this paper are as follows:
\begin{enumerate}
 \item We divide the update gradient of value distribution into two parts: (a) mean-related gradient, which incorporates the first-order term of random returns, and (b) variance-related gradient, which includes the second-order term of random returns. While standard DSAC reduces the randomness of variance-related gradient by introducing a target return clipping boundary into gradient calculation, it overlooks the randomness reduction of mean-related gradient. DSAC-T uses expected value substituting to stabilize mean-value updates, where the random target return is replaced by a more stable target Q-value, essentially the expected value of the target return. This adjustment leads to a similar update rule as non-distributional methods like SAC. Our empirical experiments show that this refinement significantly improves learning stability in practical applications.
\item In contrast to standard DSAC, which learns single value distribution, we introduce a distributional version of clipped double Q-learning, called twin value distribution learning. Specifically, DSAC-T trains two independent value distributions and employs the distribution with the lowest Q-value to calculate the gradients for both value distribution and policy function. This approach can further reduce potential overestimation bias, and moreover, it tends to introduce slight underestimation. Generally, underestimation is preferred over overestimation, as it promotes better policy performance and enhances learning stability.
\item 
During value distribution learning, standard DSAC employs a fixed clipping boundary for target return to prevent gradient explosions, but it is very sensitive to different reward scales. Additionally, the update size of the mean value of value distribution in standard DSAC is modulated by the variance, which can further exacerbate learning sensitivity in relation to reward scales. DSAC-T addresses this issue by implementing a variance-based critic gradient adjustment technique, involving substituting the fixed boundary with a variance-based value and imposing a variance-based gradient scaling weight to modulate the update size. This refinement significantly enhances the learning robustness to different reward magnitudes and markedly reduces the need for task-specific hyperparameter adjustments.
\end{enumerate}

The paper is organized as follows. Section \ref{sec:preliminaries} describes some preliminaries of RL.  Section \ref{sec:DSAC} introduces a distributional soft policy iteration framework for DSAC algorithm design. Section \ref{sec.new_version} proposes DSAC-T by incorporating three new tricks. Section \ref{sec:simulation_experiment} demonstrates the effectiveness of DSAC-T through simulations, while Section \ref{sec.application} highlights its real-world applications. Section \ref{sec.related_work} discusses the related studies, and finally, Section \ref{sec:conclusion}
 concludes the paper.

\section{Preliminaries}
\label{sec:preliminaries}

In this study, we focus on the standard setting of reinforcement learning (RL), where an agent interacts with an environment in discrete time steps \cite{li2023reinforcement}. The environment can be represented as a Markov decision process. Both state space $\mathcal{S}$ and action space $\mathcal{A}$ are assumed to be continuous. The agent receives feedback from the environment through a bounded reward signal $r(s_t,a_t)$. The state transition probability is described as $p(s_{t+1}|s_t,a_t)$, mapping a given $(s_t, a_t)$ to a probability distribution over the next state $s_{t+1}$. For simplicity, we denote the current and next state-action pairs as $(s, a)$ and $(s', a')$, respectively. The agent's behavior is defined by a stochastic policy $\pi(a_t|s_t)$, which maps a given state to a probability distribution over possible actions. The state and state-action distributions induced by $\pi$ are denoted as $\rho_{\pi}(s)$ and $\rho_{\pi}(s,a)$, respectively.

\subsection{Maximum Entropy RL}
\label{sec:max_entropy}
The aim of conventional RL is to find a policy that optimizes the expected accumulated return.  In this study, we consider an entropy-augmented objective function \cite{Haarnoja2017Soft-Q}, which supplements the reward signal with policy entropy:
\begin{equation}
\label{eq.policy_objective}
J_{\pi} = \Exp_{\substack{(s_{i\ge t},a_{i\ge t})\sim \rho_{\pi}}}\Big[\sum^{\infty}_{i=t}\gamma^{i-t} [r_i+\alpha\mathcal{H}(\pi(\cdot|s_i))]\Big],
\end{equation}
where $\gamma \in (0,1)$ is the discount factor, $\alpha$ is the temperature coefficient, and the policy entropy $\mathcal{H}$ is expressed as
\begin{equation}
\nonumber
\begin{aligned}
\mathcal{H}(\pi(\cdot|s))=\Exp_{a\sim\pi(\cdot|s)}\big[-\log\pi(a|s)\big].
\end{aligned}
\end{equation}

We denote the entropy-augmented accumulated return from $s_t$ as $G_t=\sum^{\infty}_{i=t}\gamma^{i-t} [r_i-\alpha \log\pi(a_i|s_i)]$, also referred to as soft return. The soft Q of $\pi$ is given as
\begin{equation}
\label{eq.Q_definition}
Q^{\pi}(s_t,a_t)=r_t+\gamma \Exp_{\substack{(s_{i>t},a_{i>t})\sim \rho_{\pi},\\}}[G_{t+1}],
\end{equation}
which delineates the expected soft return for choosing $a_t$ in state $s_t$ and subsequently following policy $\pi$.

The ideal policy can be found through a maximum entropy variant of policy iteration. This framework consists of two alternating stages: (a) soft policy evaluation and (b) soft policy improvement, collectively known as soft policy iteration. Provided a policy $\pi$, we can continually apply the self-consistency operator $\mathcal{T}^{\pi}$ under policy $\pi$ to learn the soft Q-value, which is depicted as follows:
\begin{equation}
\label{eq.soft_bellman}
\begin{aligned}
\mathcal{T}^{\pi}Q^{\pi}(s,a)=r+\gamma \mathbb{E}_{s'\sim p,a'\sim \pi}[Q^{\pi}(s',a')-\alpha \log\pi(a'|s')\big].
\end{aligned}
\end{equation}

On the other hand, the objective of soft policy improvement is to identify a new policy $\pi_{\rm{new}}$ that surpasses the current policy $\pi_{\rm{old}}$, such that $J_{\pi_{\rm{new}}}\ge J_{\pi_{\rm{old}}}$.
Therefore, the policy can be updated directly by maximizing the entropy-augmented objective \eqref{eq.policy_objective}, which is equivalent to
\begin{equation}
\label{eq.policy_imp}
\begin{aligned}
\pi_{\rm{new}}=\arg\max_{\pi} \Exp_{s\sim \rho_{\pi},a\sim \pi}\big[Q^{\pi_{\rm{old}}}(s,a)-\alpha \log\pi(a|s)\big].
\end{aligned}
\end{equation}

Any soft policy iteration algorithms that alternate between \eqref{eq.soft_bellman} and \eqref{eq.policy_imp} can gradually converge to the optimal maximum entropy policy. This property has been mathematically proved in \cite{Haarnoja2017Soft-Q} and \cite{Haarnoja2018SAC}.

\section{Distributional Soft Actor-Critic}
\label{sec:DSAC}

This section begins by introducing the distributional soft policy iteration (DSPI) framework, which was derived in our previous work \cite{duan2021distributional}. This framework extends maximum entropy RL into a distributional learning version. Subsequently, we outline the standard DSAC algorithm (i.e., DSACv1) that roots in this framework.

\subsection{Distributional Soft Policy Iteration}

Let us first define a random variable called soft state-action return: \begin{equation}
\nonumber
Z^{\pi}(s_t,a_t):=r_t+\gamma G_{t+1},
\end{equation}
which is a function of policy $\pi$ and state-action pair $(s_t,a_t)$. The randomness of this variable is attributed to state transition and policy.
From \eqref{eq.Q_definition}, we can observe that
\begin{equation}
\label{eq.Q_equal_exp_Z}
Q^{\pi}(s,a)=\mathbb{E}[Z^{\pi}(s,a)].
\end{equation}

One needs to model the distribution of random variable $Z^{\pi}(s,a)$. We define the function $\mathcal{Z}^{\pi}(Z^{\pi}(s,a)|s,a)$ as a mapping from $(s,a)$ to a distribution over the soft state-action return $Z^{\pi}(s,a)$. This mapping is referred to as soft state-action return distribution or simply value distribution. Relying on this definition, the distributional version of the self-consistency operator in \eqref{eq.soft_bellman} becomes
\begin{equation}\label{eq.soft_distri_bellman}
\mathcal{T}^{\pi}_{\mathcal{D}}Z(s,a) \overset{D}{=}r+\gamma( Z(s',a')-\alpha \log\pi(a'|s')),
\end{equation}
where $s'\sim p$,  $a'\sim\pi$, and $A \overset{D}{=} B$ indicates that two random variables, $A$ and $B$, have identical probability laws. 

We have proved that DSPI, which alternates between \eqref{eq.policy_imp} and \eqref{eq.soft_distri_bellman},  converges uniformly to the optimal policy. Details can be found in \cite[Appendix A]{duan2021distributional}. Equation \eqref{eq.soft_distri_bellman} defines a new random variable $\mathcal{T}^{\pi}_{\mathcal{D}}Z(s,a)$, and its distribution is denoted as $\mathcal{T}^{\pi}_{\mathcal{D}}\mathcal{Z}(\cdot|s,a)$. To solve \eqref{eq.soft_distri_bellman}, we can update the  value distribution by
\begin{equation}
\label{eq.distributional_Bellman}
\mathcal{Z}_{\rm{new}} =  \arg\min_{\mathcal{Z}}\mathop{\mathbb{E}}_{(s,a)\sim\rho_{\pi}}\big[D_{\rm KL}(\mathcal{T}^{\pi}_{\mathcal{D}}\mathcal{Z}_{\rm{old}}(\cdot|s,a),\mathcal{Z}(\cdot|s,a))\big],
\end{equation}
where $D_{\rm KL}$ is the Kullback-Leibler (KL) divergence. Note that $D_{\rm KL}$ can be replaced by other distance measures of two distributions. In fact, DSPI serves as the basic framework for developing distributional soft actor-critic algorithms, including DSACv1 in \cite{duan2021distributional} and DSAC-T in this paper.

\subsection{Standard DSAC Algorithm}

To handle continuous state and action spaces, our previous study has proposed a standard version of DSAC  (termed DSACv1), which employs neural networks as the approximators of both value function and policy function \cite{duan2021distributional}.  The value distribution and stochastic policy are parameterized as $\mathcal{Z}_{\theta}(\cdot|s,a)$ and $\pi_{\phi}(\cdot|s)$, where $\theta$ and $\phi$ are parameters. Here, we model these two parameterized functions as diagonal Gaussian, with mean and standard deviation as their outputs. Similar to most RL methods, this algorithm follows a cycle of policy evaluation (critic update) and policy improvement (actor update).

\subsubsection{Policy Evaluation}
According to \eqref{eq.distributional_Bellman}, the critic is updated by minimizing 
\begin{equation}
\label{eq.obective_general}
J_{\mathcal{Z}}(\theta)=  \mathop{\mathbb{E}}_{(s,a)\sim \mathcal{B}}\big[D_{\rm{KL}}(\mathcal{T}^{\pi_{\bar{\phi}}}_{\mathcal{D}}\mathcal{Z}_{\bar{\theta}}(\cdot|s,a),\mathcal{Z}_{\theta}(\cdot|s,a))\big],
\end{equation}
where $\mathcal{B}$ denotes the replay buffer of historical samples, while $\bar{\theta}$ and $\bar{\phi}$ are the parameters of target networks. Given that $\mathcal{T}^{\pi_{\bar{\phi}}}_{\mathcal{D}}\mathcal{Z}_{\bar{\theta}}(\cdot|s,a)$ is unknown, we develop a sample-based version of \eqref{eq.obective_general}:  
\begin{equation}
\label{eq.sample-obective_general}
J_{\mathcal{Z}}(\theta)= -\Exp_{\substack{(s,a,r,s')\sim\mathcal{B},a'\sim \pi_{\bar{\phi}},\\Z(s',a')\sim\mathcal{Z}_{\bar{\theta}}(\cdot|s',a')}}\Big[\log\mathcal{P}(y_z|\mathcal{Z}_{\theta}(\cdot|s,a))\Big],
\end{equation}
where $y_{z}$ is the target random return:
\begin{equation}
\label{eq.target_Z}
y_{z} = r+\gamma( Z(s',a')-\alpha \log\pi_{\bar{\phi}}(a'|s')).
\end{equation}

For simplicity of exposition, the subscript $(s,a,r,s')\sim\mathcal{B}$, $a'\sim \pi_{\bar{\phi}}$, $Z(s',a')\sim\mathcal{Z}_{\bar{\theta}}(\cdot|s',a')$ of $\Exp$ will be dropped if it can be inferred from the context. Given that $\mathcal{Z}_{\theta}$ is assumed to be Gaussian, it can be expressed as $\mathcal{Z}_{\theta}(\cdot|s,a)=\mathcal{N}(Q_{\theta}(s,a),\sigma_{\theta}(s,a)^2)$, where $Q_{\theta}(s,a)$ and $\sigma_{\theta}(s,a)$ are the mean and standard deviation of value distribution. Combining this with \eqref{eq.sample-obective_general}, the critic update gradient is
\begin{equation}
\label{eq:distribution_gradient}
\begin{aligned}
\nabla_{\theta}J_{\mathcal{Z}}(\theta)&=\Exp\Big[\nabla_{\theta}\frac{\left(y_{z}-Q_{\theta}(s,a)\right)^2}{2{\sigma_{\theta}(s,a)}^2}+\frac{\nabla_{\theta}\sigma_{\theta}(s,a)}{\sigma_{\theta}(s,a)}\Big]\\
&=\Exp\Big[\underbrace{-\frac{\left(y_{z}-Q_{\theta}(s,a)\right)}{{\sigma}_{\theta}(s,a)^2}\nabla_{\theta}Q_{\theta}(s,a)}_{\text{mean-related gradient}}\\
&\quad \underbrace{-\frac{\left(y_{z}-Q_{\theta}(s,a)\right)^2-{\sigma}_{\theta}(s,a)^2}{{\sigma}_{\theta}(s,a)^3}\nabla_{\theta}{\sigma}_{\theta}(s,a)}_{\text{variance-related gradient}}\Big].
\end{aligned}
\end{equation}

The critic gradient consists of two components: mean-related gradient and variance-related gradient. The difference between $y_z$ and $Q_{\theta}(s,a)$ can be viewed as a random version of the temporal difference (TD) error. The presence of squared TD error in the variance-related gradient makes the critic gradient $\nabla_{\theta}J_{\mathcal{Z}}(\theta)$ susceptible to explode as $|{\rm TD}|\rightarrow \infty$,  which may lead to learning instability. To mitigate this issue, we use a technique that clips $y_z$ in the variance-related gradient term, ensuring it stays in proximity to the mean value of the current value distribution. This technique lends stability to the learning progression of standard deviation $\sigma_{\theta}(s,a)$. The clipping function is defined as 
\begin{equation}
\label{eq.clip}
C(y_z;b):={\rm{clip}}\left(y_z,Q_{\theta}(s,a)-b,Q_{\theta}(s,a)+b\right),
\end{equation}
where $b$ is the clipping boundary. After clipping, the critic gradient \eqref{eq:distribution_gradient} becomes
\begin{equation}
\label{eq.clipped_gradient}
\begin{aligned}
\nabla_{\theta}&J_{\mathcal{Z}}(\theta)\approx\Exp\Big[-\frac{\left(y_{z}-Q_{\theta}(s,a)\right)}{{\sigma}_{\theta}(s,a)^2}\nabla_{\theta}Q_{\theta}(s,a)\\
&-\frac{\left(C(y_z;b)-Q_{\theta}(s,a)\right)^2-{\sigma}_{\theta}(s,a)^2}{{\sigma}_{\theta}(s,a)^3}\nabla_{\theta}{\sigma}_{\theta}(s,a)\Big].
\end{aligned}
\end{equation}

Moreover, the target networks employ a slow-moving update mechanism to ensure a relatively stable target distribution for critic updating.

\subsubsection{Policy Improvement}
The actor is improved by maximizing a parameterized version of \eqref{eq.policy_imp}:
\begin{equation}
\label{eq.policy_gradient}
\begin{aligned}
J_{\pi}(\phi)&=\Exp_{\substack{s\sim \mathcal{B},a\sim\pi_{\phi}}}\Big[\Exp_{Z(s,a)\sim\mathcal{Z}_{\theta}(\cdot|s,a)}[Z(s,a)]-\alpha\log(\pi_{\phi}(a|s))\Big]\\
&=\Exp_{s\sim\mathcal{B},a\sim\pi_{\phi}}[Q_{\theta}(s,a)-\alpha\log(\pi_{\phi}(a|s))],
\end{aligned}
\end{equation}
whose gradient can be trivially estimated using the reparameterization trick \cite{duan2021distributional}. 

The temperature parameter $\alpha$ plays an important role in balancing exploitation and exploration. Following similar idea in \cite{Haarnoja2018ASAC}, we update this parameter using
\begin{equation}
\label{eq.alpha_learning}
\alpha\leftarrow \alpha-\beta_{\alpha}\mathbb{E}_{s\sim\mathcal{B},a\sim \pi_{\phi}}[- \log\pi_{\phi}(a|s)-\overline{\mathcal{H}}],  
\end{equation}
where $\overline{\mathcal{H}}$ stands for the target entropy and $\beta_{\alpha}$ is the learning rate. Drawing from the discussion above, standard DSAC can be implemented by iterative updating the value distribution, policy, and temperature parameter through stochastic gradient descent (refer to \cite{duan2021distributional} for more details).

\section{DSAC with Three Refinements (DSAC-T)}
\label{sec.new_version}
The previous section introduces a standard DSAC algorithm that incorporates a distributional value function rather than an expected value function in critic and actor updates. This algorithm is practically useful, but occasionally leads to unstable learning processes in some tasks. Moreover, it requires task-specific hyperparameter tuning, which is inconvenient in fast task setups. In this section, we add three important refinements to standard DSAC, aiming to bolster learning stability and diminish sensitivity to reward scaling. These three refinements include expected value substituting, twin value distribution learning, and variance-based critic gradient adjustment. As a result, we develop an enhanced version of DSAC, named DSAC with three refinements (DSAC-T) or, alternatively, DSACv2.

\subsection{Expected value substituting}

As delineated in \eqref{eq.clipped_gradient}, DSACv1 reduces the randomness in the variance-related gradient through clipping the random target return $y_{z}$. Nonetheless, this method cannot address the high randomness in the mean-related gradient prompted by $y_{z}$. To rectify this issue, our basic idea is to replace $y_{z}$ with a steadier surrogate function.

We initially turn our attention to the target value used for Q-network updates in non-distributional methods:
\begin{equation}\label{eq.target_Q}
y_q=r+\gamma( Q_{\bar{\theta}}(s',a')-\alpha \log\pi_{\bar{\phi}}(a'|s')),
\end{equation}
where $a'\sim\pi_{\bar{\phi}}(\cdot|s')$. Compared with target Q-value $y_q$ in \eqref{eq.target_Q}, the target return $y_{z}$ in \eqref{eq.target_Z} contains more randomness due to the value distribution $\mathcal{Z}$. As indicated by the critic update gradient in  \eqref{eq.clipped_gradient}, this can lead to instability during the learning of value distribution.

From \eqref{eq.Q_equal_exp_Z}, we can show the equivalence between $y_z$ and $y_q$:
 \begin{equation}
\begin{aligned}
\label{eq.y_equal}
&\mathbb{E}_{Z(s',a')\sim\mathcal{Z}_{\bar{\theta}}(s',a')}\left[y_z\Big|_{\substack{a'\sim \pi_{\bar{\phi}},Z(s',a')\sim\mathcal{Z}_{\bar{\theta}}(\cdot|s',a')}}\right]\\
&\ =r+\gamma(Q_{\bar{\theta}}(s',a')-\alpha \log\pi_{\bar{\phi}}(a'|s'))\Big|_{a'\sim \pi_{\bar{\phi}}}\\
&\ =y_q.  
\end{aligned}
\end{equation}

Leveraging this equivalence, we can substitute the first occurrence of  $y_z$ in \eqref{eq.clipped_gradient} with $y_q$. Then, we rewrite \eqref{eq.clipped_gradient} as
\begin{equation}
\label{eq.revised_gradient}
\begin{aligned}
&\nabla_{\theta}J_{\mathcal{Z}}(\theta)\approx\Exp\Big[-\frac{\left(y_{q}-Q_{\theta}(s,a)\right)}{{\sigma}_{\theta}(s,a)^2}\nabla_{\theta}Q_{\theta}(s,a)\\
&\qquad -\frac{\left(C(y_z;b)-Q_{\theta}(s,a)\right)^2-{\sigma}_{\theta}(s,a)^2}{{\sigma}_{\theta}(s,a)^3}\nabla_{\theta}{\sigma}_{\theta}(s,a)\Big].
\end{aligned}
\end{equation}

Since $y_q$ is more certain than $y_z$, this modified critic gradient can reduce the high randomness in the mean-related gradient. Note that $y_{q}-Q_{\theta}(s,a)$ precisely represents the TD error. Consequently, the first term in  \eqref{eq.revised_gradient} is equivalent to the update gradient of Q-value in non-distributional RL methods, but scaled by an adjustment factor ${\sigma}_{\theta}(s,a)^2$. By viewing $Q_{\theta}(s,a)$ and ${\sigma}_{\theta}(s,a)$ as independent entities, the new Q-value learning mechanism delineated by \eqref{eq.revised_gradient} parallels existing RL methods like soft actor-critic (SAC) \cite{Haarnoja2018SAC}, which ensures comparable learning stability.

\subsection{Twin value distribution learning}
The second refinement is a distributional variation of clipped double Q-learning \cite{Fujimoto2018TD3}, called twin value distribution learning. Specifically, we parameterize two value distributions, characterized by parameters $\theta_1$ and $\theta_2$, which are trained independently. The value distribution with the smaller mean value is chosen to construct critic and actor gradients. For critic updating, we define the index of the chosen value distribution as 
\begin{equation}
\bar{i} := \arg\min_{i=1,2} Q_{\bar{\theta}_i}(s',a')|_{a'\sim \pi_{\bar{\phi}}(\cdot|s')}.
\end{equation}

Subsequently, we use $\bar{\theta}_{\bar{i}}$ to evaluate the target return in \eqref{eq.target_Z} and the target Q-value in \eqref{eq.target_Q}. The expressions for these target evaluations are shown as follows:
\begin{equation}
\begin{aligned}
\label{eq.target_Z_min}
y^{\min}_{z} &= r+\gamma( Z(s',a')-\alpha \log\pi_{\bar{\phi}}(a'|s'))\big|_{Z(s',a')\sim \mathcal{Z}_{\bar{\theta}_{\bar{i}}}(\cdot|s',a')},\\
y^{\min}_q&=r+\gamma( Q_{\bar{\theta}_{\bar{i}}}(s',a')-\alpha \log\pi_{\bar{\phi}}(a'|s')).
\end{aligned}
\end{equation}

By inserting \eqref{eq.target_Z_min} into \eqref{eq.revised_gradient}, it follows that 
\begin{equation}
\label{eq.final_return_gradient}
\begin{aligned}
&\nabla_{\theta_i}J_{\mathcal{Z}}(\theta_i)\approx\Exp\Big[-\frac{\left(y^{\min}_{q}-Q_{\theta_i}(s,a)\right)}{{\sigma}_{\theta_i}(s,a)^2}\nabla_{\theta_i}Q_{\theta_i}(s,a)\\
&\  -\frac{\left(C(y^{\min}_z;b)-Q_{\theta_i}(s,a)\right)^2-{\sigma}_{\theta_i}(s,a)^2}{{\sigma}_{\theta_i}(s,a)^3}\nabla_{\theta_i}{\sigma}_{\theta_i}(s,a)\Big].
\end{aligned}
\end{equation}

In a similar manner, the actor objective undergoes a revision of twin value distributions:
\begin{equation}
\label{eq.actor_objetive_2}
\begin{aligned}
J_{\pi}(\phi)=\Exp_{s\sim\mathcal{B},a\sim\pi_{\phi}}[\min_{i=1,2}Q_{\theta_i}(s,a)-\alpha\log(\pi_{\phi}(a|s))].
\end{aligned}
\end{equation}  

In contrast to the single value distribution used in DSACv1, DSAC-T employs the minimization of twin value distributions to shape the target distribution in the critic update. This approach is able to further mitigate overestimation bias, and moreover, it has the tendency to yield slight underestimation. It is important to highlight that minor underestimation is generally more desirable than overestimation. This is because overestimated action values can propagate during learning, whereas underestimated actions typically get sidestepped by the policy, preventing their values from propagating. Additionally, the underestimation of Q-values can serve as a performance lower bound for policy optimization, which is helpful to improve learning stability. 

\subsection{Variance-based critic gradient adjustment}

As per \eqref{eq.clip}, DSACv1 adopts a fixed clipping boundary to prevent gradient explosions. The choice of this clipping boundary is of significant importance, as a smaller value can impact the accuracy of variance learning, while a larger value may lead to a huge gradient norm. An improper selection of clipping boundaries may severely hinder learning performance. It is important to recognize that the mean and variance of value distribution have a direct relationship with reward magnitudes. This indicates that different reward scales typically correspond to distinct optimal clipping boundary designs. Moreover, reward magnitudes can vary not only across diverse tasks but also evolve over time as the policy improves during training. Therefore, DSACv1 often needs manual adjustment of reward scales for each specific task, which is a non-trivial job in itself.

To alleviate this sensitivity to reward scaling, our refined version automates the clipping boundary determination by 
\begin{equation}
\label{eq.adaptive_boundary}
b=\xi\Exp_{\substack{(s,a)\sim\mathcal{B}}}\left[{\sigma}_{\theta}(s,a)\right],
\end{equation}
where $\xi$ is a constant parameter that controls the clipping range. In this setup, the boundary can adapt to varying reward magnitudes across varied tasks and training phases. Compared to the direct adjustment of $b$, selecting $\xi$ is more straightforward, as \eqref{eq.adaptive_boundary} inherently correlates with reward magnitude. Typically, we can set $\xi=3$ following the three-sigma rule. While this refinement is straightforward, it is remarkably effective, eliminating the necessity for task-specific hyperparameter fine-tuning. 

Furthermore, as previously noted, the update size of the mean value $Q_{\theta}(s,a)$ in DSAC is modulated by the variance, as shown in the denominator of \eqref{eq.final_return_gradient}. This variance sensitivity distinguishes DSAC from non-distributional RL algorithms, which may also lead to sensitivity in learning with respect to reward scales. To address this, we introduce a gradient scaling weight $\omega$, leading to the scaled objective function:
\begin{equation}
\label{eq.scale_obective_general}
J^{\rm scale}_{\mathcal{Z}}(\theta)=  \omega\mathop{\mathbb{E}}_{(s,a)\sim \mathcal{B}}\big[D_{\rm{KL}}(\mathcal{T}^{\pi_{\bar{\phi}}}_{\mathcal{D}}\mathcal{Z}_{\bar{\theta}}(\cdot|s,a),\mathcal{Z}_{\theta}(\cdot|s,a))\big],
\end{equation}
where 
\begin{equation}
\label{eq.adaptive_scale}
\omega=\Exp_{\substack{(s,a)\sim\mathcal{B}}}\left[{\sigma}_{\theta}(s,a)^2\right].
\end{equation}

By using a moving average technique for both $\omega$ and $b$, the corresponding gradient for each value distribution is
\begin{equation}
\label{eq.scale_return_gradient}
\begin{aligned}
&\nabla_{\theta_i}J^{\rm scale}_{\mathcal{Z}}(\theta_i)\approx\\
&\quad(\omega_i+\epsilon_\omega)\Exp\Big[-\frac{\left(y^{\min}_{q}-Q_{\theta_i}(s,a)\right)}{{\sigma}_{\theta_i}(s,a)^2+\epsilon}\nabla_{\theta_i}Q_{\theta_i}(s,a)\\
&\  -\frac{\left(C(y^{\min}_z,b_i)-Q_{\theta_i}(s,a)\right)^2-{\sigma}_{\theta_i}(s,a)^2}{{\sigma}_{\theta_i}(s,a)^3+\epsilon}\nabla_{\theta_i}{\sigma}_{\theta_i}(s,a)\Big],
\end{aligned}
\end{equation}
where $\epsilon$ and $\epsilon_\omega$ are small positive numbers. The $\epsilon$ is introduced to prevent gradient explosion when ${\sigma}_{\theta_i}(s,a)\rightarrow 0$, while $\epsilon_\omega$ is used to prevent gradient disappearance as $\omega_i\rightarrow 0$.

The update rules for $b_i$ and $\omega_i$, with $\tau$ as the synchronization rate, are detailed as
\begin{equation}
\begin{aligned}
&b_i \leftarrow \tau\xi\Exp_{\substack{(s,a)\sim\mathcal{B}}}\left[{\sigma}_{\theta_i}(s,a)\right]+(1-\tau)b_i,\\
&\omega_i\leftarrow\tau\Exp_{\substack{(s,a)\sim\mathcal{B}}}\left[{\sigma}_{\theta_i}(s,a)^2\right]+(1-\tau)\omega_i.
\end{aligned}
\end{equation}

Finally, DSAC-T is detailed in Algorithm \ref{alg:DSAC-T}. It is important to note that both DSAC-T and DSACv1 are built upon the DSPI framework, which has been proven to converge uniformly to the optimal solution (see \cite[Appendix A]{duan2021distributional} for details). Since DSAC-T introduces refinements only at the algorithmic level, its theoretical convergence properties remain the same as those of DSACv1. 
\begin{algorithm}[!htb]
\caption{DSAC-T}
\label{alg:DSAC-T}
\begin{algorithmic}
\STATE Input:  $\theta_1$, $\theta_2$, $\phi$, $\alpha$, $\beta_{\mathcal{Z}}$, $\beta_{\pi}$, $\beta_{\alpha}$, $\tau$ 
\STATE Initialize target networks: $\bar{\theta}_1\leftarrow\theta_1$, $\bar{\theta}_2\leftarrow\theta_2$, $\bar{\phi}\leftarrow\phi$

\FOR{each iteration}
\FOR{each sampling step}
\STATE Calculate action $a\sim\pi_{\phi}(a|s)$
\STATE Get reward $r$ and new state $s'$
\STATE Store samples $(s,a,r,s')$ in buffer $\mathcal{B}$
\ENDFOR
\FOR{each update step}
\STATE Sample data from $\mathcal{B}$
\STATE Update critic using $\theta \leftarrow \theta - \beta_{\mathcal{Z}}\nabla_{\theta} J^{\rm scale}_{\mathcal{Z}}(\theta)$
\STATE Update actor using $\phi \leftarrow \phi + \beta_{\pi}\nabla_{\phi} J_{\pi}(\phi)$
\STATE Update temperature using \eqref{eq.alpha_learning}
\STATE Update target networks using 
\STATE $\quad\bar{\theta} \leftarrow  \tau\theta+(1-\tau)\bar{\theta}, \;  
\bar{\phi} \leftarrow  \tau\phi+(1-\tau)\bar{\phi}$
\ENDFOR
\ENDFOR
\end{algorithmic}
\end{algorithm}

\section{Experiments}
\label{sec:simulation_experiment}

\begin{figure}
\centering
\captionsetup[subfigure]{justification=centering}
\subfloat[\label{subFig:envhumanoid}]
{\includegraphics[width = 0.12\textwidth, height = 0.12\textwidth]{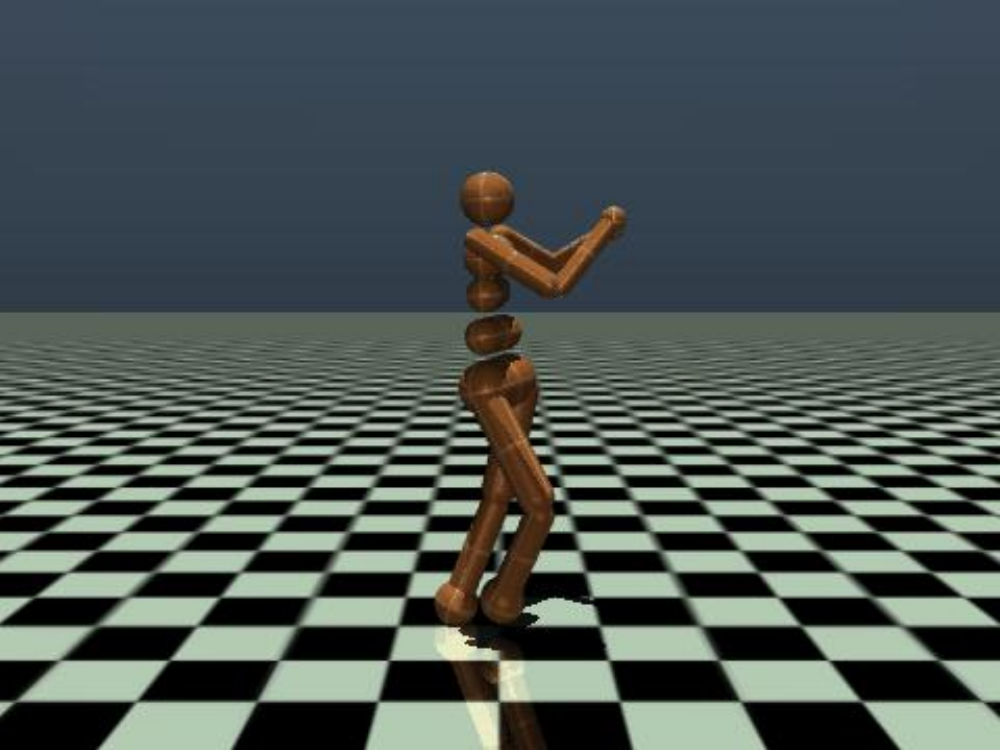}} 
\subfloat[\label{subFig:envant}]
{\includegraphics[width = 0.12\textwidth, height = 0.12\textwidth]{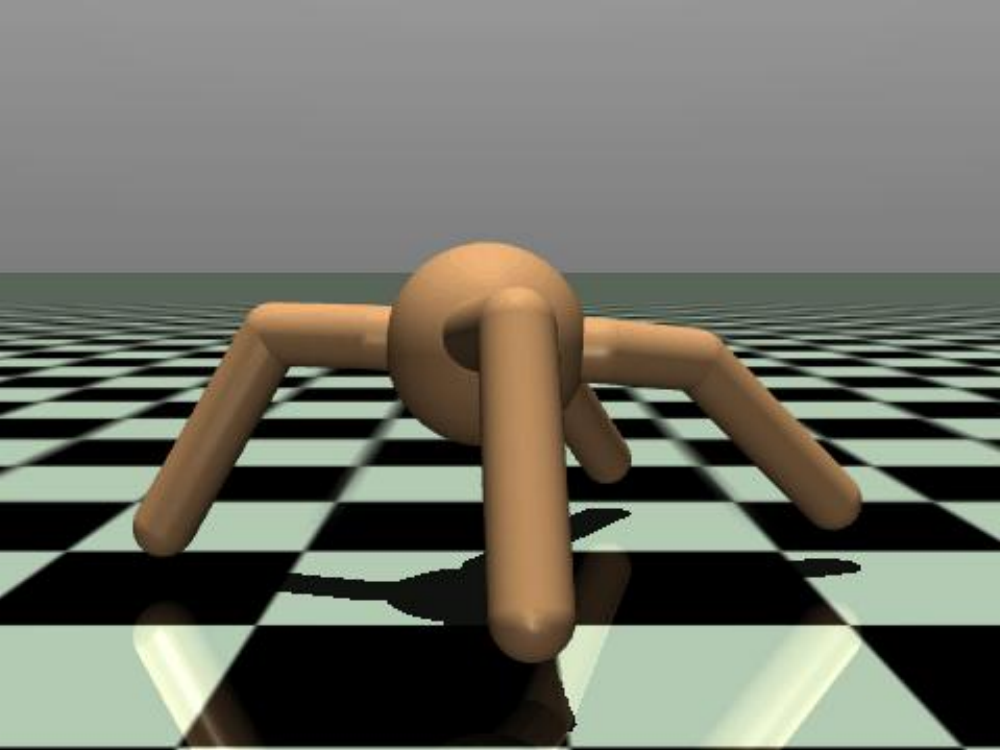}}
\subfloat[\label{subFig:envhalfcheetah}]
{\includegraphics[width = 0.12\textwidth, height = 0.12\textwidth]{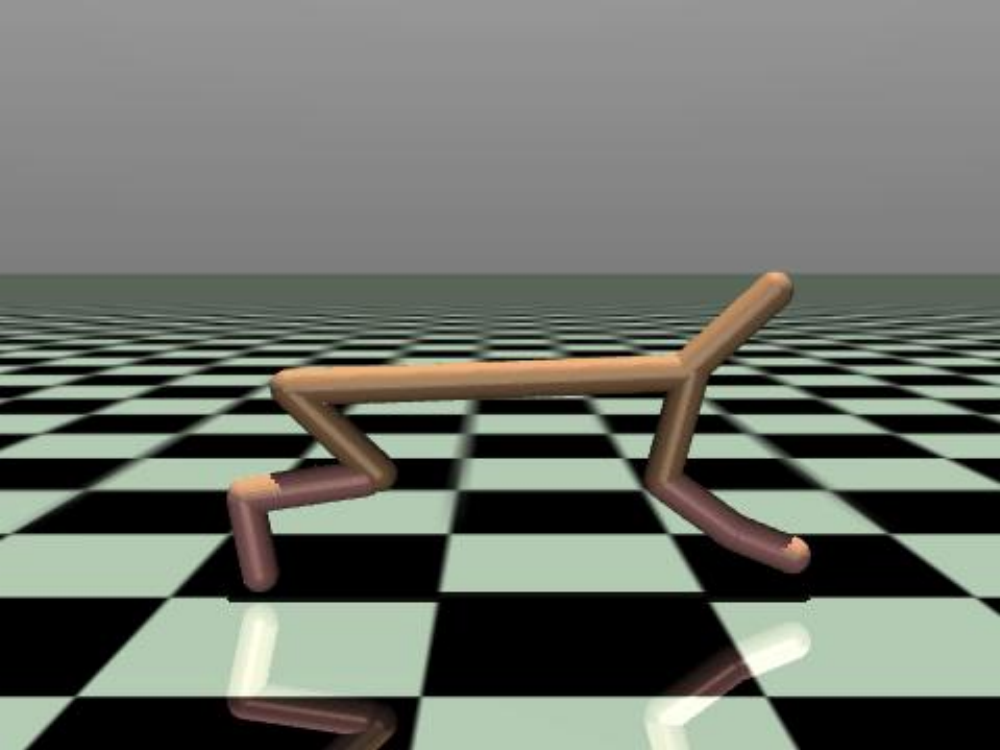}} 
\subfloat[\label{subFig:envwalker2d}]
{\includegraphics[width = 0.12\textwidth, height = 0.12\textwidth]{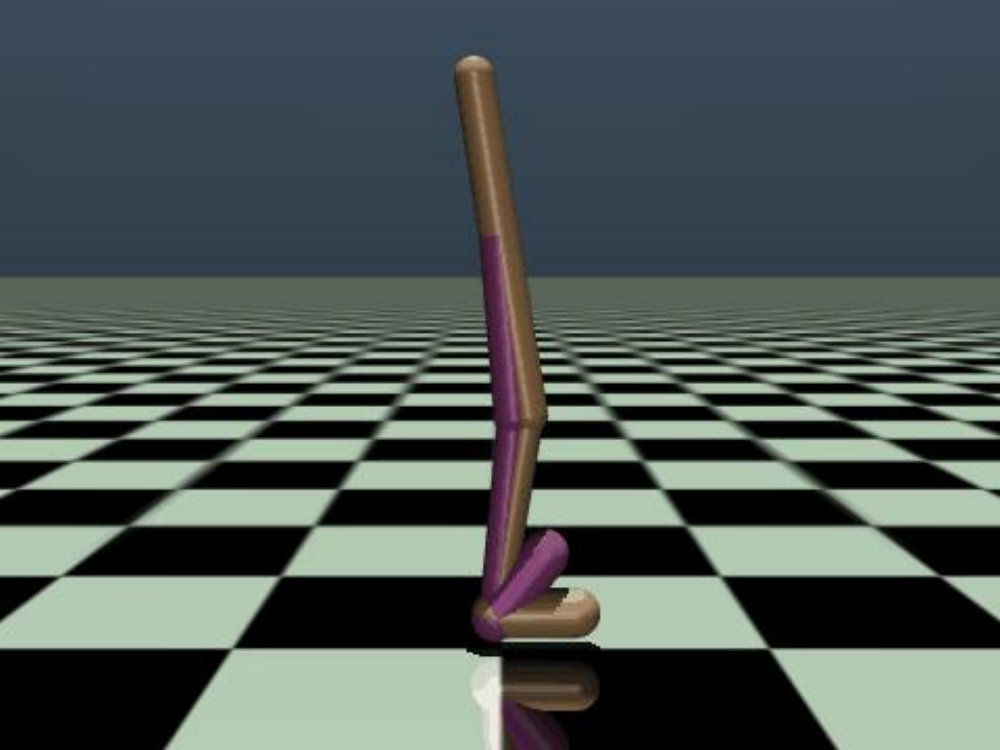}} 
\vspace{0.1em}
\subfloat[\label{subFig:envidp2d}]
{\includegraphics[width = 0.12\textwidth, height = 0.12\textwidth]{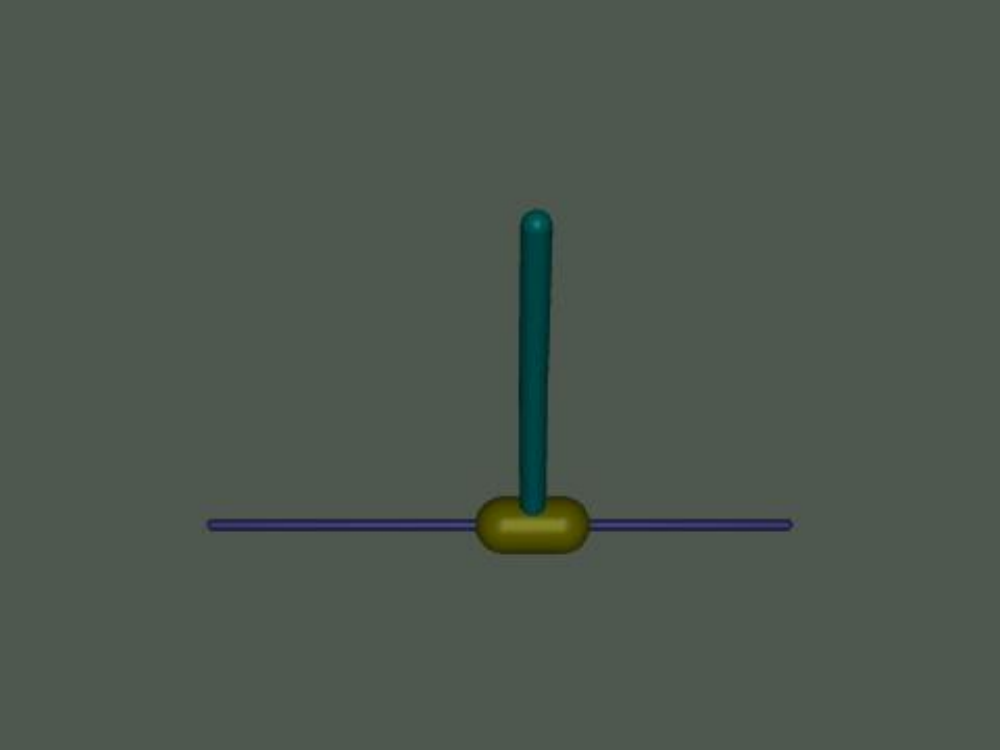}}
\subfloat[\label{subFig:envhopper}]
{\includegraphics[width = 0.12\textwidth, height = 0.12\textwidth]{figures/hopper.pdf}}
\subfloat[\label{subFig:envpusher}]
{\includegraphics[width = 0.12\textwidth, height = 0.12\textwidth]{figures/pusher.pdf}}
\subfloat[\label{subFig:envreacher}]
{\includegraphics[width = 0.12\textwidth, height = 0.12\textwidth]{figures/reacher.pdf}} 
\vspace{0.1em}
\subfloat[\label{subFig:envswimmer}]
{\includegraphics[width = 0.12\textwidth, height = 0.12\textwidth]{figures/swimmer.pdf}} 
\subfloat[\label{subFig:envbipedalwalker}]
{\includegraphics[width = 0.12\textwidth, height = 0.12\textwidth]{figures/bipedalwalker.pdf}}
\subfloat[\label{subFig:envbipedalwalker_hard}]
{\includegraphics[width = 0.12\textwidth, height = 0.12\textwidth]{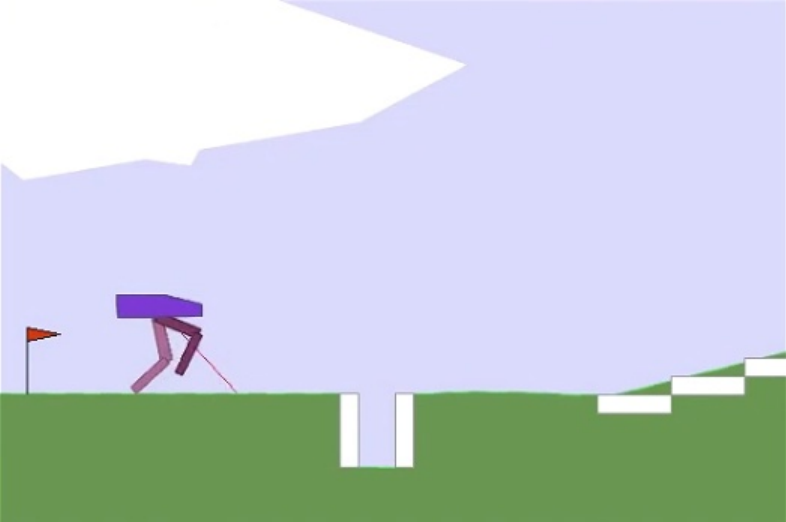}}
\subfloat[\label{subFig:envcarracing}]
{\includegraphics[width = 0.12\textwidth, height = 0.12\textwidth]{figures/carracing.pdf}}

\caption{Benchmarks. 
    (a) Humanoid-v3: \((s \times a) \in \mathbb{R}^{376} \times \mathbb{R}^{17}\). 
    (b) Ant-v3: \((s \times a) \in \mathbb{R}^{111} \times \mathbb{R}^{8}\).
    (c) HalfCheetah-v3: \((s \times a) \in \mathbb{R}^{17} \times \mathbb{R}^{6}\).
    (d) Walker2d-v3: \((s \times a) \in \mathbb{R}^{17} \times \mathbb{R}^{6}\).
    (e) InvertedDoublePendulum-v2: \((s \times a) \in \mathbb{R}^{6} \times \mathbb{R}^{1}\).
    (f) Hopper-v3: \((s \times a) \in \mathbb{R}^{11} \times \mathbb{R}^{3}\).
    (g) Pusher-v2: \((s \times a) \in \mathbb{R}^{23} \times \mathbb{R}^{7}\).
    (h) Reacher-v2: \((s \times a) \in \mathbb{R}^{11} \times \mathbb{R}^{2}\).
    (i) Swimmer-v3: \((s \times a) \in \mathbb{R}^{8} \times \mathbb{R}^{2}\).
    (j) BipedalWalker-v3: \((s \times a) \in \mathbb{R}^{24} \times \mathbb{R}^{4}\).
    (k) BipedalWalker-hardcore-v3: \((s \times a) \in \mathbb{R}^{24} \times \mathbb{R}^{4}\).
    (l) CarRacing-v1:  \((s \times a) \in \mathbb{R}^{96\times 96 \times 3} \times \mathbb{R}^{2}\) (image-input).
    }

\label{f:envs}
\end{figure}

We assess the performance of DSAC-T by conducting a series of continuous control tasks facilitated through the OpenAI Gym interface. The benchmark tasks utilized in this study are depicted in Fig. \ref{f:envs}, including 11 vector-input-based control tasks (Humanoid, Ant, HalfCheetah, Walker2d, InvertedDoublePendulum, Hopper, Pusher, Reacher, Swimmer,  Bipedalwalker, and Bipedalwalker-hardcore) and one image-input-based control task (CarRacing). For the vector-input-based tasks, the state consists of the physical positions and velocities of the robot's joints, while the action corresponds to the torque applied to these joints. The agent earns positive rewards for maintaining good posture and moving toward the goal and is penalized for failing to complete the control task or applying excessive torque. For the image-input-based task (CarRacing), the state consists of three-channel pixel values, and the action controls the car's acceleration/braking and steering. A small negative reward is assigned at each step to encourage the car to finish the race quickly, while positive rewards are given when the car reaches checkpoints. Detailed instructions for the experimental settings can be found on the OpenAI Gym website \cite{brockman2016openaigym}. All baseline algorithms used are accessible in GOPS \cite{wang2023gops}, an open-source RL solver developed with PyTorch. 

\begin{figure*}[t]
\centering
\captionsetup[subfigure]{justification=centering}
\subfloat[Humanoid-v3\label{subFig:humanoid}]
{\includegraphics[width = 0.2325\textwidth]{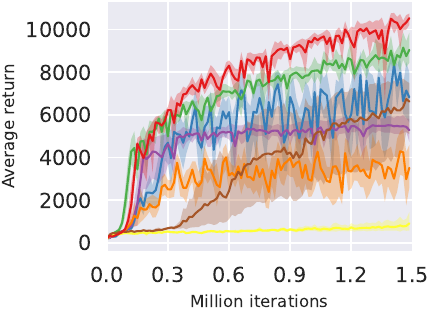}} \quad
\subfloat[Ant-v3\label{subFig:ant}]
{\includegraphics[width = 0.2325\textwidth]{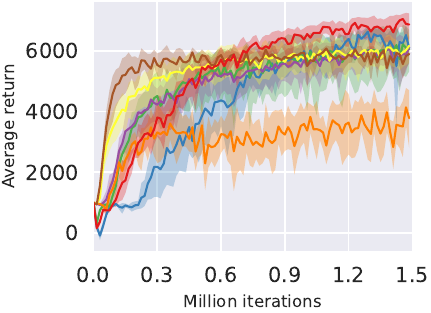}} \quad
\subfloat[Halfcheetah-v3\label{subFig:halfcheetah}]
{\includegraphics[width = 0.2325\textwidth]{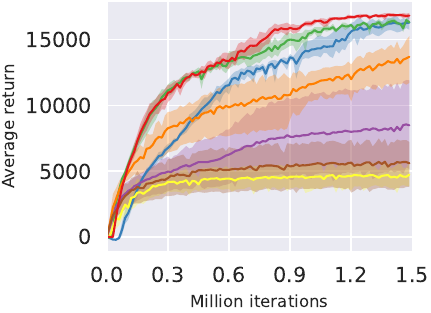}} \quad
\subfloat[Walker2d-v3\label{subFig:walker2d}]
{\includegraphics[width = 0.2325\textwidth]{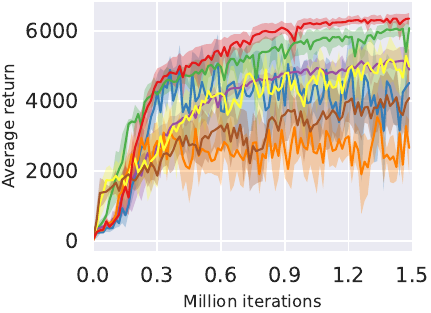}} \quad
\vspace{0.5em}
\subfloat[I.D.Pendulum-v2\label{subFig:idp}]
{\includegraphics[width = 0.2325\textwidth]{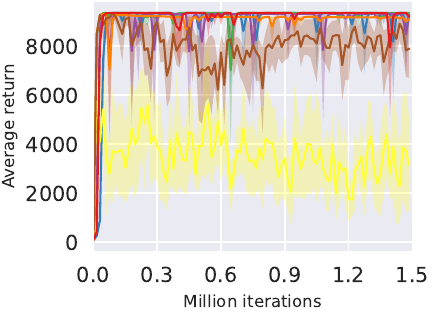}} \quad
\subfloat[Hopper-v3\label{subFig:hopper}]
{\includegraphics[width = 0.2325\textwidth]{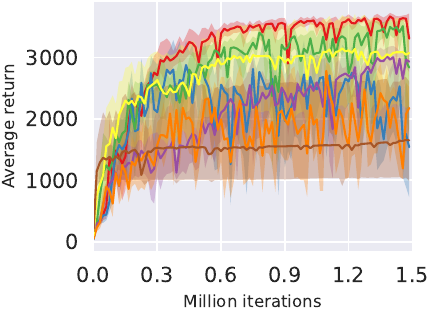}} \quad
\subfloat[Pusher-v2\label{subFig:pusher}]
{\includegraphics[width = 0.2325\textwidth]{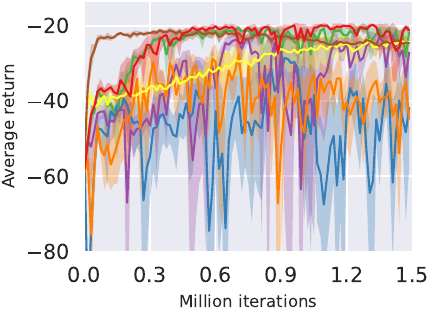}} \quad
\subfloat[Reacher-v2\label{subFig:reacher}]
{\includegraphics[width = 0.2325\textwidth]{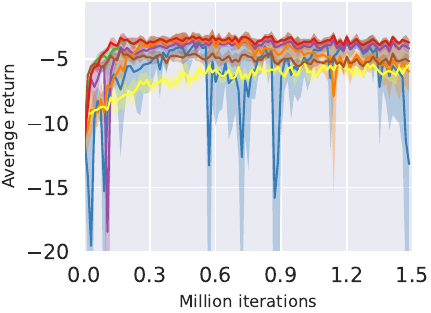}} \quad
\vspace{0.5em}
\subfloat[Swimmer-v3\label{subFig:swimmer}]
{\includegraphics[width = 0.2325\textwidth]{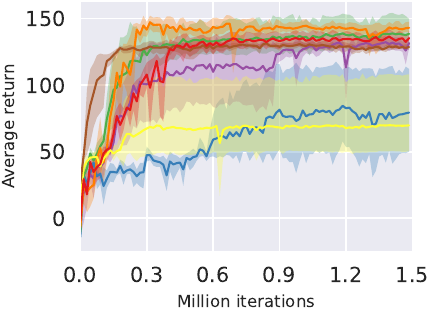}} \quad
\subfloat[Bipedalwalker-v3\label{subFig:bipedalwalker}]
{\includegraphics[width = 0.2325\textwidth]{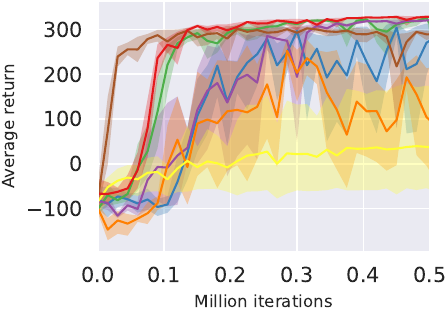}} \quad
\subfloat[B.walker-hardcore-v3\label{subFig:bipedalwalker_hardcore}]
{\includegraphics[width = 0.2325\textwidth]{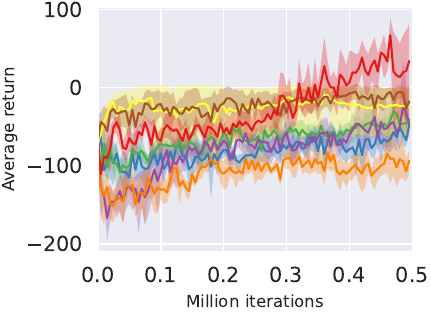}} \quad
 \subfloat[CarRacing-v1\label{subFig:carracing}]
 {\includegraphics[width = 0.2325\textwidth]{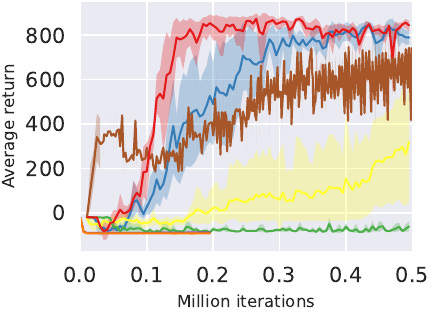}} \quad
 \subfloat
{\includegraphics[width = 0.9\textwidth]{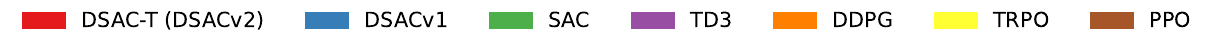}} \quad
\caption{Training curves on benchmarks. The solid lines correspond to mean and shaded regions
correspond to 95\% confidence interval over five runs. The iteration of PPO and TRPO is counted by the number of network updates.}
\label{f:benchmeark}
\end{figure*}

\subsection{Baselines}
\label{sec:baseline}
Our algorithm is evaluated against well-known model-free algorithms. These include deep deterministic policy gradient (DDPG) \cite{lillicrap2015DDPG}, trust region policy optimization (TRPO) \cite{schulman2015TRPO}, proximal policy optimization (PPO) \cite{schulman2017PPO}, twin delayed deep deterministic policy gradient (TD3) \cite{Fujimoto2018TD3}, and soft actor-critic (SAC) \cite{Haarnoja2018ASAC}. These baselines have been widely tested and employed across a range of demanding domains. By comparing with these algorithms, we aim to provide an objective evaluation of DSAC-T. We also draw comparisons between DSAC-T and DSACv1, where DSAC-T employs an adaptive clipping boundary with $\xi=3$ and DSACv1 utilizes a fixed value of $b=20$.

To maintain fairness in comparison, our DSAC-T algorithm is also implemented in GOPS, ensuring an identical modular architecture. For the value distribution and stochastic policy, we employ a diagonal Gaussian distribution. Specifically, each neural network projects the input states to the mean and standard deviation. The Adam optimization method is employed for all parameter updates. All algorithms, including baselines and DSAC-T, follow a similar network architecture and use equivalent hyperparameters. Specifically, for tasks with vector inputs, we implement a multi-layer perceptron architecture for both the actor and critic networks. Each of these networks comprises three hidden layers, with each layer containing 256 units and using GELU activations. For tasks involving image inputs (specifically CarRacing), an extra encoding network is integrated. This network is tailored for embedding three-channel image observations into a 256-dimensional hidden state, comprising six sequential convolutional layers with RELU activation functions. These layers are characterized by convolutional kernels of sizes $[4,3,3,3,3,3]$ and strides $[2,2,2,2,1,1]$. Basic hyperparameters are provided in Table \ref{t.hyper}, and the training files containing full hyperparameter details are accessible at \href{https://github.com/Intelligent-Driving-Laboratory/GOPS}{\emph{https://github.com/Intelligent-Driving-Laboratory/GOPS}}.

\begin{table}[!htp]
\captionsetup{justification=centering,labelsep=newline,font=small}
\captionsetup{justification=centering,labelsep=newline,font={small,sc}}
\caption{Detailed hyperparameters.}
\label{t.hyper}
\begin{tabular}{lc}
\toprule
Hyperparameters & Value \\
\hline
\emph{Shared} & \\
\quad Optimizer &  Adam ($\beta_{1}=0.9, \beta_{2}=0.999$)\\
\quad Actor learning rate & $1{\rm{e-}}4 $\\
\quad Critic learning rate & $1{\rm{e-}}4 $\\
\quad Discount factor ($\gamma$) & 0.99 \\
\quad & (0.999 specifically for Hopper)\\
\quad Policy update interval & 2\\
\quad Target smoothing coefficient ($\tau$) & 0.005\\
\quad Reward scale & 1\\
\quad Random seed set& [12345,22345,32345,42345,52345]\\
\hline
\emph{Maximum-entropy framework} &\\ 
\quad  Learning rate of $\alpha$ &  $3{\rm{e-}}4 $ \\
\quad  Expected entropy ($\overline{\mathcal{H}}$) &  $\overline{\mathcal{H}}=-{\rm{dim}}(\mathcal{A})$ \\
\hline
\emph{Deterministic policy} &\\ 
\quad Exploration noise&  $\epsilon \sim \mathcal{N}(0,0.1^2)$\\
\hline
\emph{Off-policy} &\\ 
\quad Replay buffer warm size & $1\times10^4$\\
\quad Replay buffer size & $1\times10^6$\\
\quad Samples collected per iteration &  20 \\
\hline
\emph{On-policy} &\\ 
\quad Sample batch size &  2000 \\
\quad Replay batch size &  2000 \\
\quad GAE factor ($\lambda$)  & 0.95\\
\hline
\emph{DSAC-T} &\\ 
\quad $\zeta$ in \eqref{eq.adaptive_boundary} &  3 \\
\quad  $\epsilon$ and $\epsilon_{\omega}$ in \eqref{eq.scale_return_gradient} &  0.1 \\
\bottomrule
\end{tabular}
\end{table}

\subsection{Results}

We conducted five independent training executions for each experiment, using five different random seeds (listed in Table \ref{t.hyper}) that were consistent across all algorithms and benchmarks. Learning curves and policy performance are presented in Fig. \ref{f:benchmeark} and Table \ref{t.benchmark}, respectively. Our results reveal that DSAC-T surpasses (at least matches) the performance of all baseline algorithms across all benchmark tasks. Taking Humanoid-v3 as an example, compared with SAC, TD3, PPO, DDPG, and TRPO, our algorithm shows relative improvements of 16.0\%, 92.3\%, 57.7\%, 104.7\%, and 1022.2\%, respectively. These results suggest that DSAC-T sets a new standard of performance for model-free RL algorithms. Moreover, compared to its predecessor (DSACv1), this new version (DSAC-T) has achieved substantial enhancements in both learning stability and final outcomes.

\begin{table*}[!htb]
 \centering
\captionsetup{justification=centering,labelsep=newline,font={small,sc}}
    \caption{Average final return. Computed as the mean of the highest return values observed in the final 10\% of iteration steps per run, with an evaluation interval of 15,000 iterations. The maximum value for each task is bolded. $\pm$ corresponds to standard deviation over five runs.}
\label{t.benchmark}
\begin{tabular}{c c c c c c c c}
  \toprule
  Task & DSAC-T & DSACv1 & SAC & TD3 & DDPG & TRPO & PPO \\ \hline
  Humanoid-v3 & \textbf{10829} $\pm$ \textbf{243}& 9074 $\pm$ 286 & 9335 $\pm$ 695 & 5631 $\pm$ 435 & 5291 $\pm$ 662 & 965 $\pm$ 555 & 6869 $\pm$ 1563 \\
  Ant-v3 & \textbf{7086} $\pm$ \textbf{261}& 6862 $\pm$ 53& 6427 $\pm$ 804 & 6184 $\pm$ 486 & 4549 $\pm$ 788 & 6203 $\pm$ 578 & 6156 $\pm$ 185 \\
  Halfcheetah-v3 & \textbf{17025} $\pm$ \textbf{157}& 16541 $\pm$ 514 & 16573 $\pm$ 224 & 8632 $\pm$ 4041 & 13970 $\pm$ 2083 & 4785 $\pm$ 967 & 5789 $\pm$ 2200 \\
  Walker2d-v3 & \textbf{6424} $\pm$ \textbf{147}& 5413 $\pm$ 865 & 6200 $\pm$ 263 & 5237 $\pm$ 335 & 4095 $\pm$ 68 & 5502 $\pm$ 593 & 4831 $\pm$ 637 \\
  Inverteddoublependulum-v2 & \textbf{9360} $\pm$ \textbf{0} & 9359 $\pm$ 1 & \textbf{9360} $\pm$ \textbf{0} & 9347 $\pm$ 15 & 9183 $\pm$ 9 & 6259 $\pm$ 2065 & 9356 $\pm$ 2 \\
  Hopper-v3 & \textbf{3688} $\pm$ \textbf{61}&  3098$\pm$223& 3551 $\pm$ 131 & 3176 $\pm$ 120 & 2933 $\pm$ 167 & 3138 $\pm$ 870 & 1679 $\pm$ 1000 \\
  Pusher-v2 & \textbf{-19} $\pm$ \textbf{1} &  -26$\pm$1& -20 $\pm$ 0 & -21 $\pm$ 1 & -30 $\pm$ 6 & -23 $\pm$ 2 & -23 $\pm$ 1 \\
  Reacher-v2 & \textbf{-3} $\pm$ \textbf{0}&  -4$\pm$2& \textbf{-3} $\pm$ \textbf{0} & \textbf{-3} $\pm$ \textbf{0} & -4 $\pm$ 1 & -5 $\pm$ 1 & -4 $\pm$ 0 \\
  Swimmer-v3 & 138 $\pm$ 6 &  84$\pm$36& 140 $\pm$ 14 & 134 $\pm$ 5 & \textbf{146} $\pm$ \textbf{4} & 70 $\pm$ 38& 130 $\pm$ 2 \\
  Bipedalwalker-v3 & \textbf{330} $\pm$ \textbf{3} &  319$\pm$2 & 327 $\pm$ 1 & 322 $\pm$ 2 & 306 $\pm$ 9 & 40 $\pm$ 137& 303 $\pm$ 3\\
  Bipedalwalker-hardcore-v3 & \textbf{77} $\pm$ \textbf{31} &  -31$\pm$9 & -14 $\pm$ 8 & -1 $\pm$ 52 & -55 $\pm$ 24 & -7 $\pm$ 18& -1 $\pm$ 9\\
  CarRacing-v1 & \textbf{903} $\pm$ \textbf{7}&  890$\pm$12& -12 $\pm$ 8& -89$\pm$ 5& -93$\pm$ 0& 363 $\pm$ 323& 776$\pm$ 98\\
  \bottomrule
\end{tabular}
\end{table*}

\begin{table*}[!htb]  
\captionsetup{justification=justified,labelsep=newline,font={small,sc}}
      \centering
      \caption{
Average value estimation bias over five runs. This bias is computed using (${\text{estimate Q-value}-\text{true Q-value}}$), where the true Q value is assessed based on the discounted accumulation of sampled rewards, with the entropy reward term added for maximum entropy-based algorithms. The best value is in bold. The superscript $^\star$ indicates superior estimation accuracy of DSAC-T over off-policy baselines including SAC, TD3, and DDPG. Meanwhile, $^\dagger$ denotes superior estimation accuracy of DSAC-T over on-policy baselines like TRPO and PPO.  When biases are comparable, underestimation is more favorable than overestimation.
}
    \label{t.bias}
    \begin{tabular}{c c c c c c c c}
        \toprule
        Task & DSAC-T & DSACv1 & SAC & TD3 & DDPG & TRPO & PPO \\ \hline
        Humanoid-v3 & \textbf{-42.29}$^\star$$^\dagger$ & 207.25 & -81.69 & -226.05 & 48.80 & 18.72 & 17.28 \\
        Ant-v3 & \textbf{-10.55}$^\star$$^\dagger$ & 39.04 & -25.31 & -327.33 & 89.36 & 13.36 & 8.37 \\
        Halfcheetah-v3 & 23.95$^\dagger$ & 73.02 & \textbf{-4.82} & -341.37 & 31.81 & 603.82 & 95.20 \\
        Walker2d-v3 & \textbf{-0.79}$^\star$$^\dagger$ & 72.94 & -5.49 & -60.05 & 128.54 & 5.90 & 1.80 \\
Inverteddoublependulum-v2 & 2.60$^\star$ & 77.12 & 5.68 & -560.99 & 57632.34 & 3.32 & \textbf{1.57} \\
        Hopper-v3 & \textbf{-7.00}$^\star$$^\dagger$ & 2171.96 & 252.12 & -718.11 & 547666.18 & 275.47 & 245.66 \\
        Pusher-v2 & -6.83$^\dagger$ & \textbf{-4.71} & -7.02 & -10.76 & 1.19 & -10.35 & -8.68 \\
        Reacher-v2 & -5.46 & -5.38 & -5.44 & -10.13 & \textbf{-0.28} & -6.87 & -4.99 \\
        Swimmer-v3 & 0.60 & 1.90 & \textbf{-0.07} & -1.00 & 0.10 & 0.15 & 0.02 \\
        Bipedalwalker-v3 & -3.80$^\star$ & 0.17 & -3.98 & -9.78 & 10.99 & \textbf{-0.72} & -3.12 \\
        Bipedalwalker-hardcore-v3 & \textbf{-30.36}$^\star$$^\dagger$ & 126.30 & 79.60 & -159.34 & 617.22 & -125.21 & -101.69 \\
        CarRacing-v1 & -117.07 & 26.13 & \textbf{-0.38} & 1.09 & 1.10 & -2.93 & 26.95 \\
        \bottomrule
      \end{tabular}
\end{table*}

Table \ref{t.bias} showcases the value estimation bias for each algorithm. While both DSACv1 and DDPG utilize a single critic (excluding the target critic) in an off-policy manner, DSACv1 exhibits lower overestimation bias overall. This suggests that value distribution learning can partly counteract overestimation issues. By incorporating the twin value distribution learning technique, DSAC-T further reduces overestimation bias, leading to a minor underestimation in certain benchmarks. As a result, DSAC-T achieves enhanced learning stability compared to DSACv1. Guided by the principle that underestimation is preferred over overestimation when biases are of similar magnitude, the estimation accuracy of DSAC-T either surpasses or at least aligns with all off-policy baselines across most benchmarks. Even when compared to on-policy baselines like PPO and TRPO, DSAC-T consistently demonstrates a significant advantage in estimation accuracy across many benchmarks. We also assessed the computational efficiency of DSAC-T by comparing the average time taken per 1,000 iterations against other off-policy baselines in Humanoid-v3. The results are as follows: DSAC-T (35.51s), DSACv1 (29.00s), SAC (35.02s), TD3 (31.41s), and DDPG (25.20s). While DSAC-T takes slightly longer than the others, the significant performance improvements it delivers justify the additional computational cost.

\begin{figure}[!htb]
\centering
\captionsetup[subfigure]{justification=centering}
\subfloat[Humanoid-v3\label{subFig:humanoid_comp}]
{\includegraphics[width = 0.2325\textwidth]{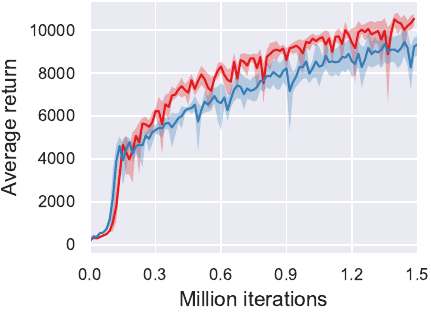}} \quad
\subfloat[Ant-v3\label{subFig:ant_comp}]
{\includegraphics[width = 0.2325\textwidth]{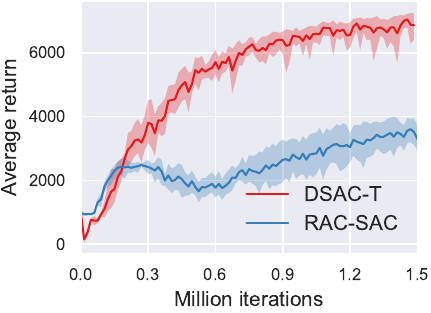}} \quad
\caption{Training curve comparison between DSAC-T and RAC-SAC. The solid lines correspond to mean and shaded regions correspond to 95\% confidence interval over five runs.}
\label{f:compare}
\end{figure}

In addition to the mainstream baselines, we compared DSAC-T with a recent overestimation suppression method, realistic actor critic-soft actor critic (RAC-SAC) \cite{sicen_realistic_2023}, across the two most complex benchmarks: Humanoid and Ant (using consistent random seeds and hyperparameters). RAC-SAC learns an ensemble of value functions with varying confidence bounds, and the variance of these ensemble Q-values is used as a regularization term in the critic update target to reduce overestimation. The results show that DSAC-T achieves superior estimation accuracy in both Humanoid (estimation bias: -42.29 vs. 97.78) and Ant (estimation bias: -10.55 vs. -18.82). Given DSAC-T’s significantly higher performance, as shown in Fig. \ref{f:compare}, it shows substantial improvement in Q-value estimation accuracy when considering relative estimation accuracy (absolute bias/Q-value).

\subsection{Ablation Studies}
Subsequently, we carry out ablation studies to evaluate the impact of individual refinement within DSAC-T. 
\subsubsection{Learning stability} 
\begin{figure}[!htb]
\centering{\includegraphics[width = 0.465\textwidth]{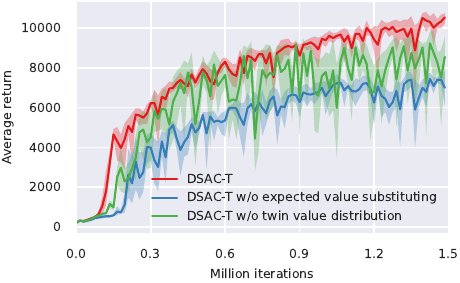}}
\caption{Training curves in Humanoid-v3 over five runs, comparing ablation over expected value substituting and twin value distribution learning.}
\label{fig_ablation1}
\end{figure}

As displayed in Fig. \ref{fig_ablation1}, DSAC-T outshines its variants, DSAC-T without expected value substituting (which calculates the critic gradient using \eqref{eq.clipped_gradient} instead of \eqref{eq.revised_gradient}) and DSAC-T with a single value distribution, in both learning stability and overall performance. This confirms the significant contributions of the expected value substitution technique, defined by \eqref{eq.revised_gradient}, and the twin value distribution learning technique to learning stability and efficacy. Moreover, the ascending trajectory of the training curve reveals that the inclusion of the expected value substituting refinement also accelerates learning by reducing gradient randomness.

\subsubsection{Sensitivity to reward scaling} 
\begin{figure}[!htb]
\centering{\includegraphics[width = 0.465\textwidth]{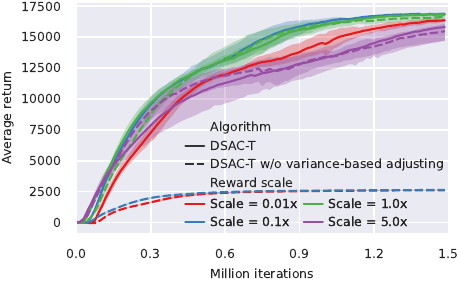}}
\caption{Training curves in Halfcheetah-v3 over five runs with different reward scales, comparing ablation over variance-based critic gradient adjustment.}
\label{fig_ablation2}
\end{figure}

Fig. \ref{fig_ablation2} illustrates the comparative performance of DSAC-T and DSAC-T without variance-based critic gradient adjustment (using a fixed boundary value $b=20$) across varied reward scales. DSAC-T maintains similar performance across diverse reward magnitudes, in contrast to DSAC-T without gradient adjusting, which exhibits significant sensitivity to the alterations in reward scales. Notably, it struggles to learn an effective policy under lower reward scales, specifically at 0.01 and 0.1. This comparison 
 emphasizes the efficacy of incorporating a variance-based clipping boundary and gradient scaling weight in DSAC-T, which significantly reduces the need for adjusting hyperparameters to suit specific tasks.

\section{Real-world Applications}
\label{sec.application}

\begin{figure}[!htb]
\centering{\includegraphics[width = 0.47\textwidth]{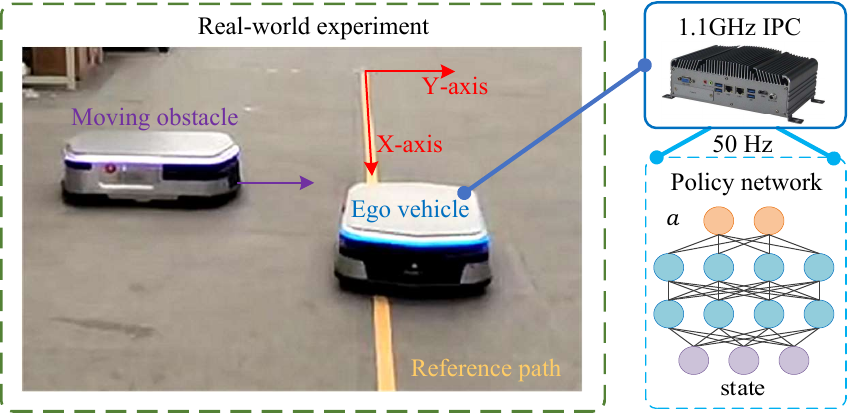}}
\caption{Mobile robot trajectory tracking and collision avoidance experiment setup. The robot is powered by an Intel(R) Pentium(R) N4200 CPU for computation, with the policy network operating at a control frequency of 50Hz.}
\label{f:exp_scenario}
\end{figure}

This section describes the practical application of DSAC-T in controlling mobile robots, demonstrating its viability for real-world scenarios. We utilized the Geekplus M200 mobile robot as the experimental platform, tasked with precisely following a reference path (as illustrated in Fig. \ref{f:exp_scenario}) with a desired speed of 0.28 m/s, all while avoiding collisions with obstacles that could emerge from any direction at a steady speed. The control actions defined for the robot are its acceleration and angular acceleration. The states include the robot's relative position and angle to both the reference path and the obstacle, in addition to its current velocity and angular velocity. The reward function aims to minimize errors in position and velocity alignment with the target path, discourage large action magnitudes, and heavily penalize collisions with obstacles. The neural network architecture and the specific hyperparameters chosen for the algorithm in this real-world application are consistent with those detailed in Section \ref{sec:simulation_experiment}.

\begin{figure}[!htb]
\centering\includegraphics[width = 0.45\textwidth]{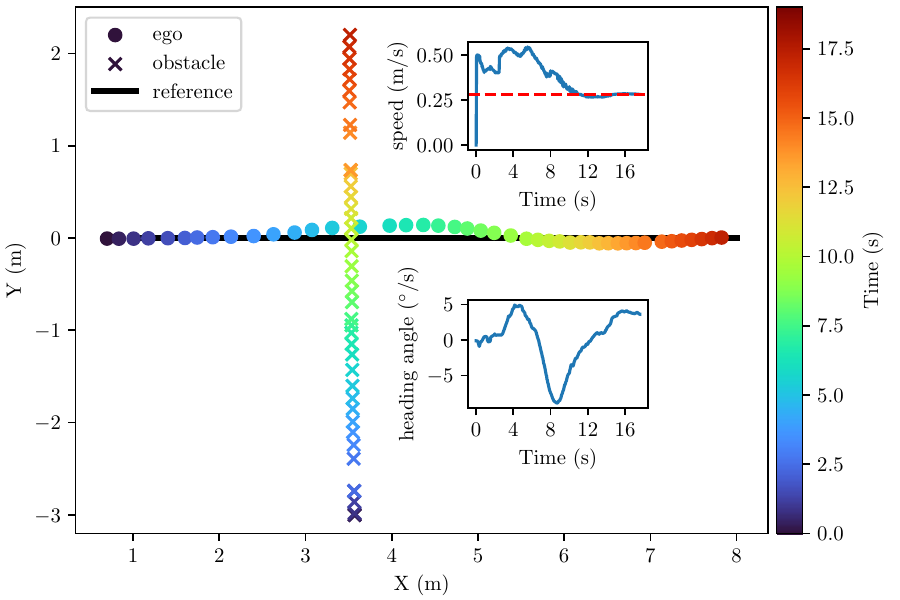} 
\caption{State curves during driving. Points are uniformly spaced in time, making point density indicative of object speed. }
\label{f:DSACT_trjectory}
\end{figure}

\begin{figure*}[!htb]
\centering
\captionsetup[subfigure]{justification=centering}
\subfloat[t=0]
{\label{f:keyframe_0}\includegraphics[width = 0.183\textwidth,height = 0.12\textwidth]{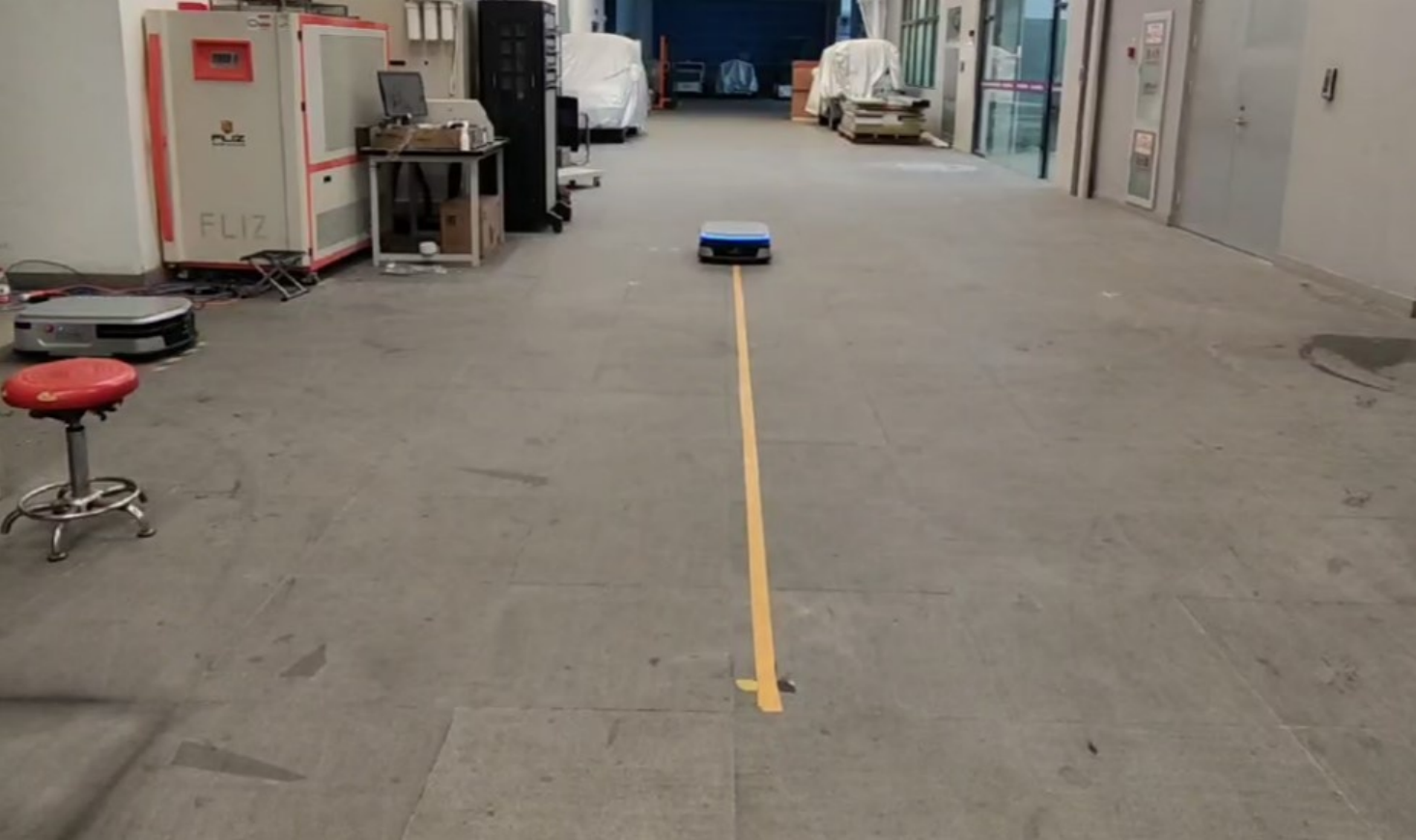}} 
\subfloat[t=4s]
{\label{f:keyframe_4}\includegraphics[width = 0.183\textwidth,height = 0.12\textwidth]{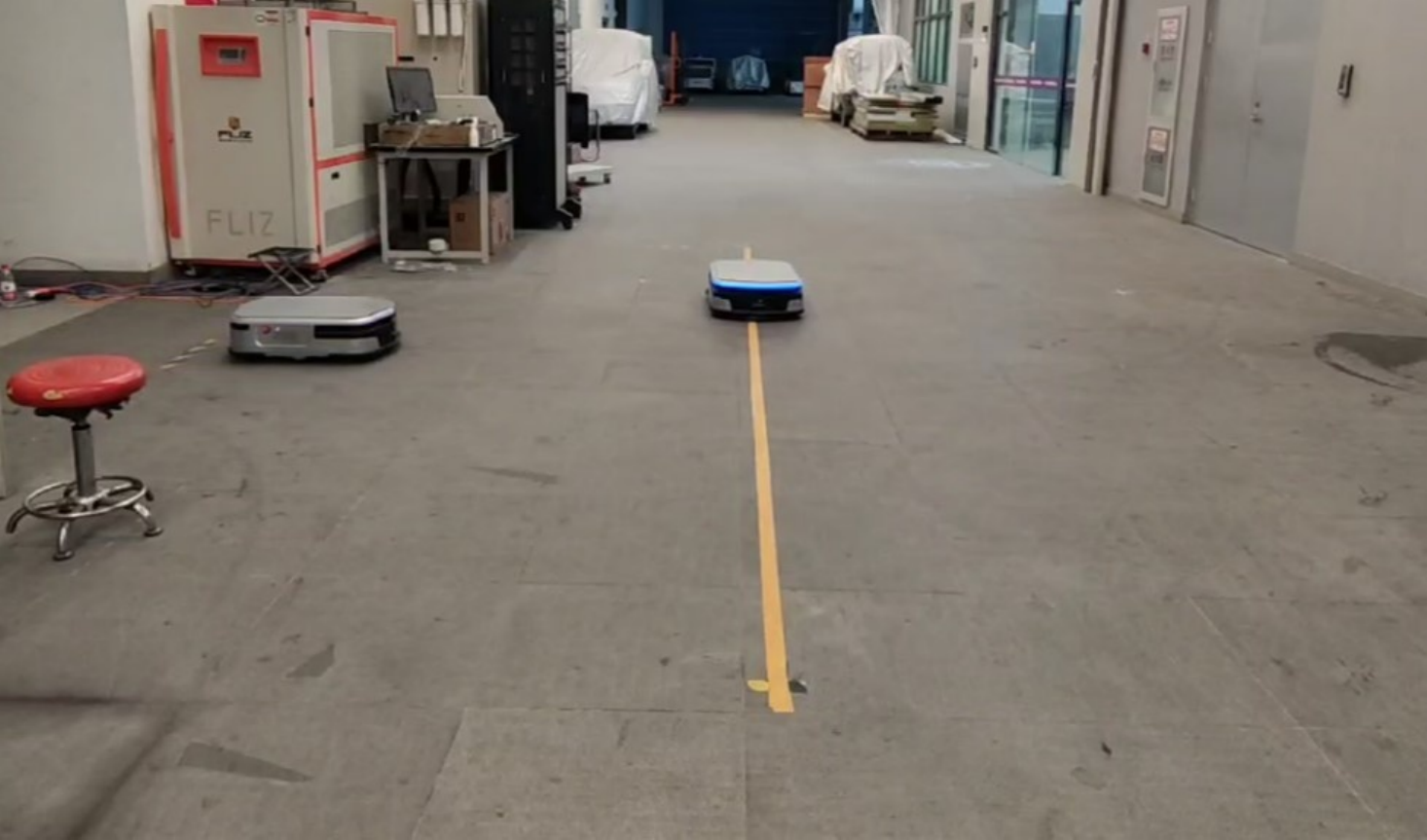}} 
\subfloat[t=8s]
{\label{f:keyframe_8}\includegraphics[width = 0.183\textwidth,height = 0.12\textwidth]{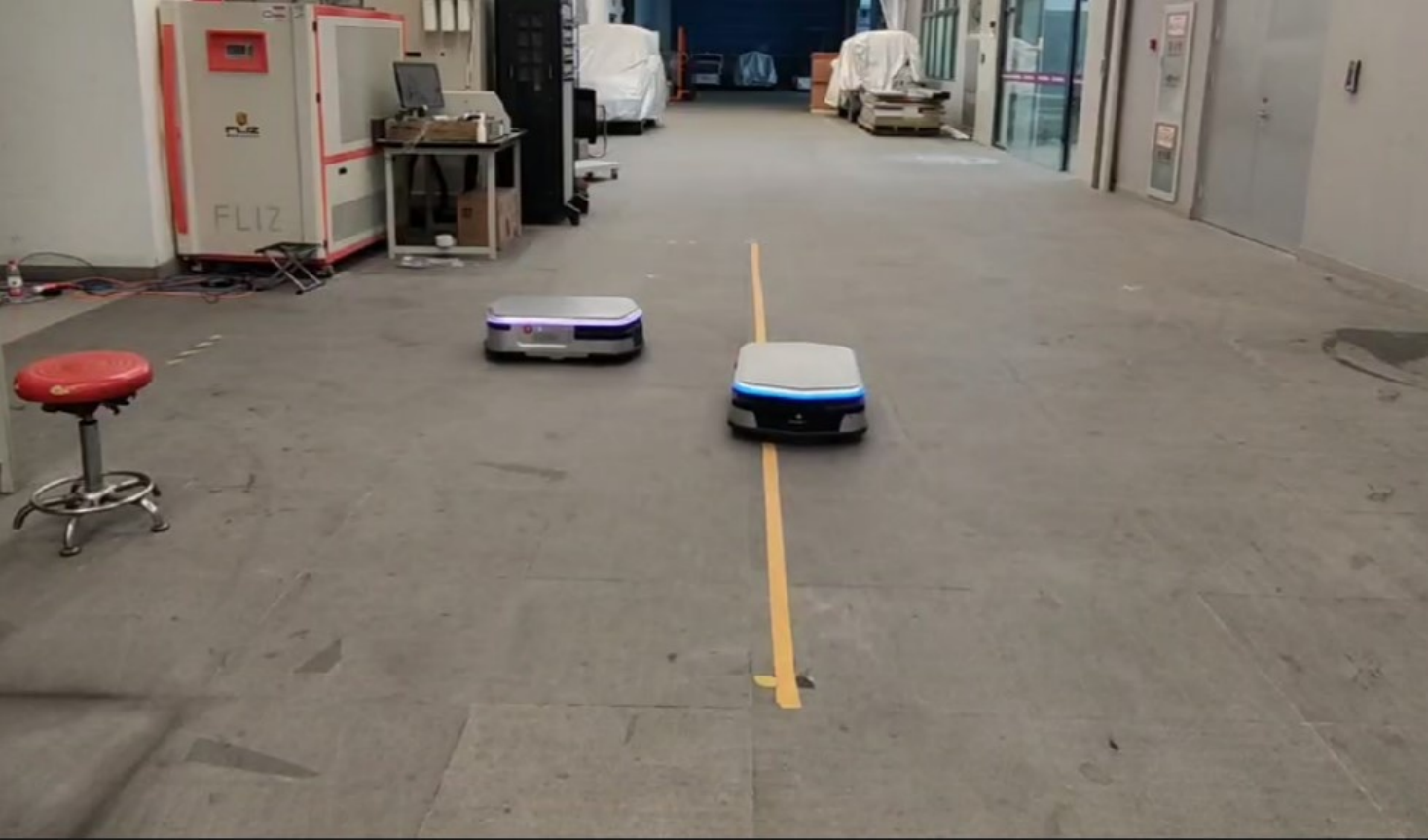}} 
\subfloat[t=12s]
{\label{f:keyframe_12}\includegraphics[width = 0.183\textwidth,height = 0.12\textwidth]{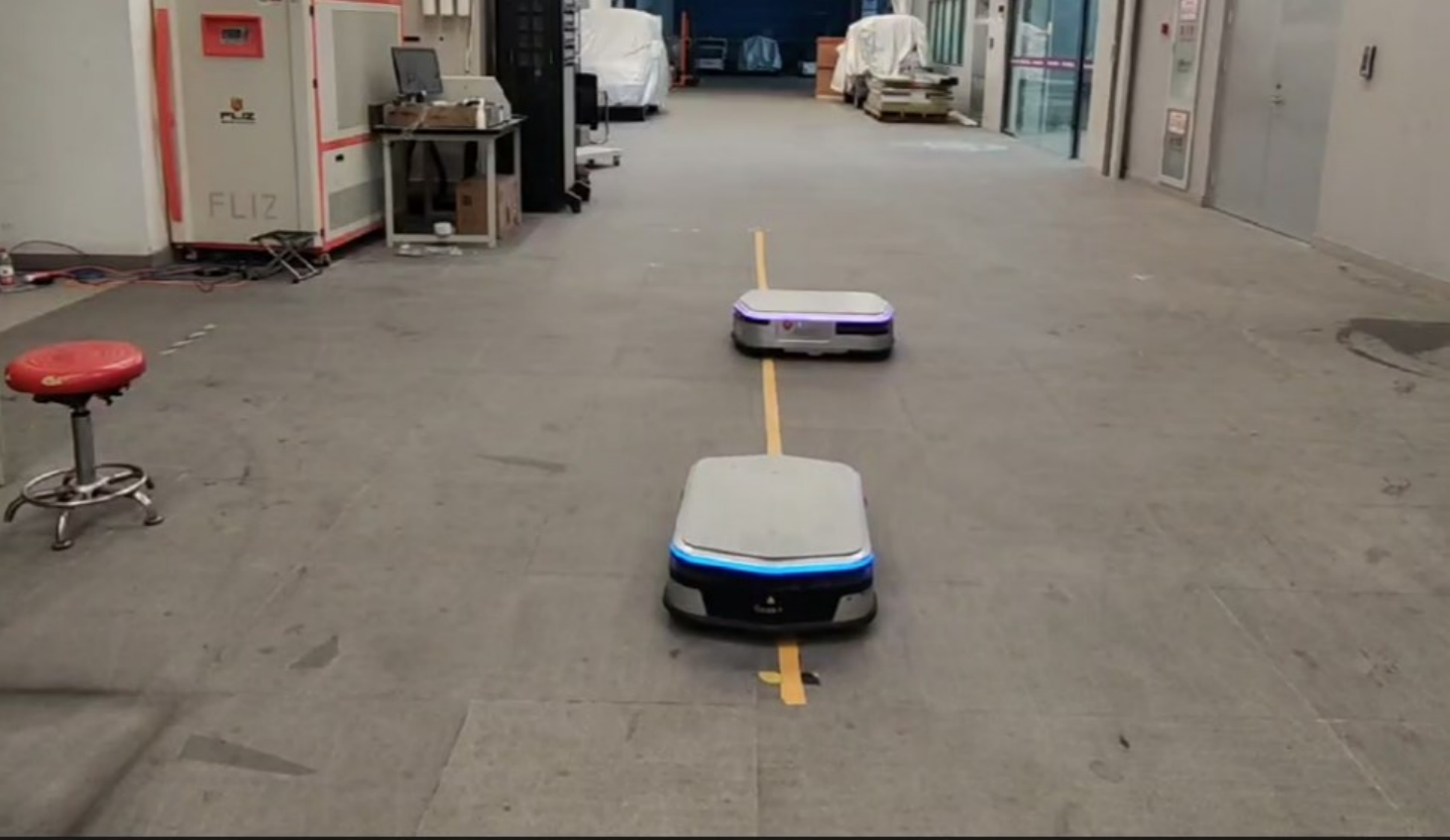}} 
\subfloat[t=16s]
{\label{f:keyframe_16}\includegraphics[width = 0.183\textwidth,height = 0.12\textwidth]{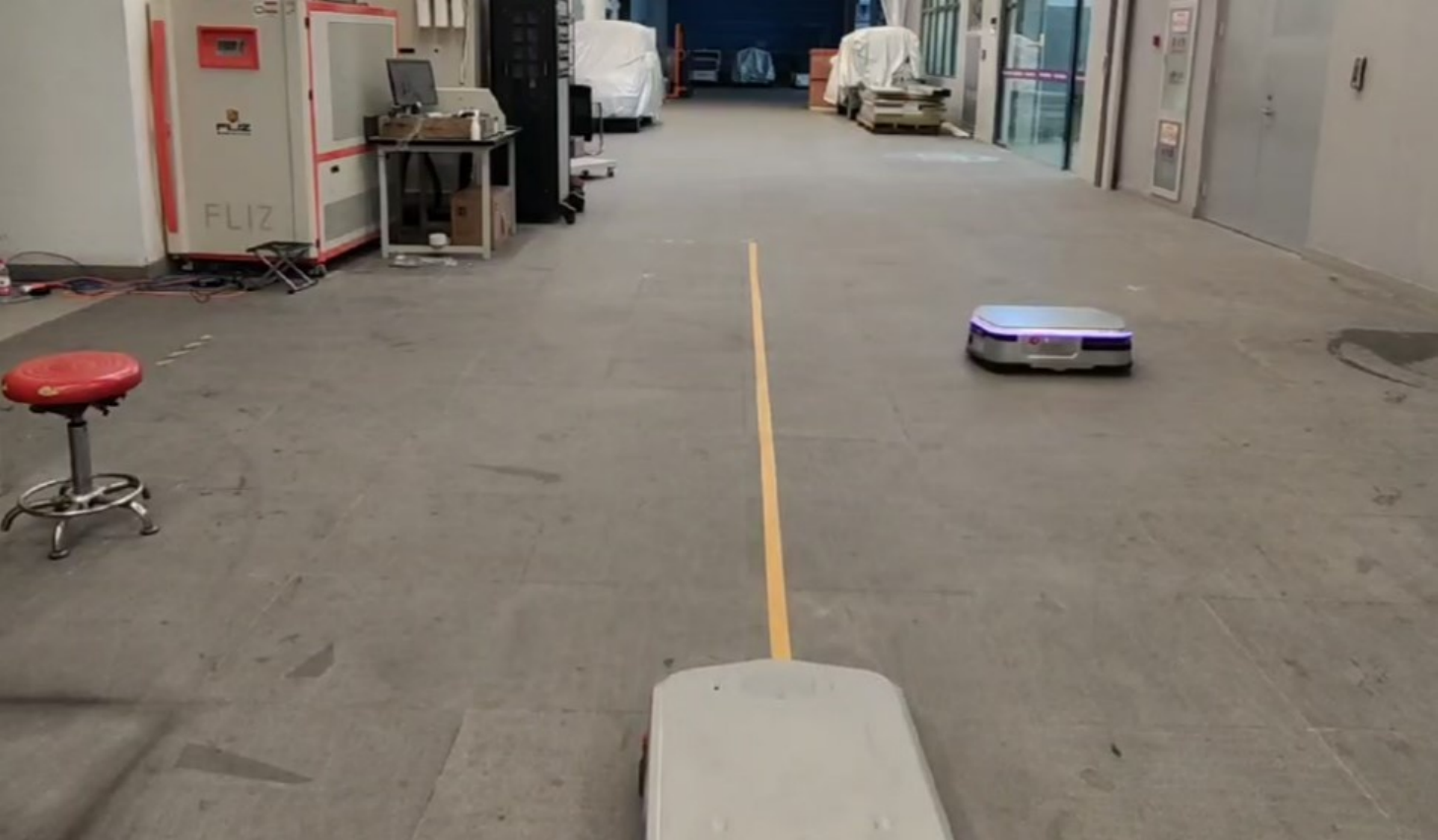}} 
\caption{Driving process snapshot.}
\label{f:key_frames}
\end{figure*}

The driving policy is learned from interactions within a simulated environment that utilizes the robot's kinematic model and is subsequently deployed on the mobile robot for practical applications. Despite the inherent discrepancies between simulation and actual environments, the trained policy effectively enables the robot to complete tasks involving path tracking and obstacle avoidance. A representative driving sequence is visualized in Fig. \ref{f:DSACT_trjectory}, with crucial moments captured in Fig. \ref{f:key_frames}. In this sequence, the robot initiates from a stationary state. To circumvent collisions and avoid halting, it opts to swiftly accelerate to 0.5 m/s and travels around this speed. Concurrently, as an obstacle moves towards the robot, it executes a left turn to evade the obstacle. Once past the potential collision zone, the robot realigns with the planned path and reduces its speed to the target velocity of 0.28 m/s. This demonstrates the potential of DSAC-T for practical control problems.

\section{Related Work}
\label{sec.related_work}
The concept of distributional RL, which involves modeling the distribution of returns with its expectation representing the value function, was initially introduced in \cite{bellemare2017C51}. Since then, numerous distributional RL algorithms have emerged, catering to both discrete control settings \cite{dabney2018quantileregre,davney2018quantilenet,yang2019fully,rowland2019statistics,mavrin2019distributional} and continuous control settings \cite{barth-maron2018D4PG,tessler2019distributional}. Building on the insights gained from distributional RL research, Dabney \emph{et al.} \cite{dabney2020dopamine-basedRL} observed from mouse experiments that the brain represents potential future rewards not as a single mean but as a probability distribution. 

Prevalent distributional RL studies typically shaped the value distribution utilizing either a discrete probability density function or a quantile function, resulting in a scenario where the mean value is not directly accessible but must be approximated based on a weighted average of finite quantiles. When the objective is to find a policy that maximizes the expected return, the evaluation accuracy of the mean value (i.e., Q-value) becomes crucial, even more so than the overall accuracy of the distribution. In this study, we directly learn a continuous probability density function for random returns, predicated on a Gaussian assumption, with Q-value as a direct output of the parameterized critic function. The theoretical findings presented in \cite{duan2021distributional} demonstrate that this approach effectively mitigates Q-value overestimation. Furthermore, the introduction of twin value distribution learning has further reduced the overestimation bias, facilitating improved policy performance. Our ongoing work involves conducting a comprehensive analysis of various value distribution learning methods to continue advancing the field of distributional RL.

Similar to non-distributional RL, DSAC-T updates the policy by maximizing the Q-value. However, value distributions provide richer information for policy learning than Q-values alone \cite{bellemare_distributional_2023}. For instance, the variance of the value distribution can be incorporated into the objective \eqref{eq.policy_gradient} to promote risk-sensitive policy learning \cite{ryg2020minmaxDSAC,chandak_universal_2021, brown_bayesian_2020}. Additionally, the uncertainty captured by the value distribution can be used to guide exploration, enhancing policy performance \cite{willian_estimating_2019,xiao_MDSAC_2024}. Leveraging value distribution effectively for policy improvement is a promising area for future research.

\section{Conclusion}
\label{sec:conclusion}
In this study, we present an enhanced version of the distributional soft actor-critic algorithm, called DSAC with three refinements (DSAC-T or DSACv2), designed to address issues of learning instability, sensitivity to reward scaling, and to further improve the accuracy of Q-value estimation. These refinements encompass expected value substituting, twin value distribution learning, and variance-based critic gradient adjustment. Specifically, we achieve improved stability by replacing the random target return term of the mean-related gradient with a more stable target Q-value. This is coupled with the deployment of a twin value distribution learning scheme. Empirical results substantiate the claim that these two refinements significantly improve learning stability, thereby boosting policy performance. To counter the reward scaling sensitivity, the third refinement introduces an adaptive variance-based clipping boundary for random target returns, along with an adaptive critic gradient scaling weight. Experiment results demonstrate that DSAC-T exhibits a noteworthy performance enhancement when compared to mainstream model-free RL methods, including SAC, TD3, DDPG, PPO, and TRPO. Additionally, the successful application of DSAC-T in controlling a real-world wheeled robot establishes it as a promising solution for real-world challenges.

Regarding the limitations, in the current version of DSAC-T, we utilize a unimodal Gaussian distribution to approximate both the value distribution and the stochastic policy. However, in tasks with multiple goals and complex dynamics, the true distributions may exhibit multimodal characteristics. This limitation in representational capacity may hinder the agent's exploration efficiency and its ability to learn the optimal policy. Future work will explore new distributional RL methods by incorporating multimodal approximation techniques, such as diffusion models, to overcome this limitation.







\ifCLASSOPTIONcaptionsoff
  \newpage
\fi

\bibliographystyle{ieeetr}
\bibliography{reference}

\end{document}